\documentclass{article} % DO NOT CHANGE THIS
\usepackage{aaai25}  % DO NOT CHANGE THIS
\usepackage{times}  % DO NOT CHANGE THIS
\usepackage{helvet}  % DO NOT CHANGE THIS
\usepackage{courier}  % DO NOT CHANGE THIS
\usepackage[hyphens]{url}  % DO NOT CHANGE THIS
\usepackage{graphicx} % DO NOT CHANGE THIS

\usepackage{latexsym}
\usepackage{amssymb}
\usepackage{amsmath}
\usepackage{amsthm}
\usepackage{booktabs}
\usepackage{enumitem}

\usepackage{color}
\usepackage{listings}
\usepackage[most]{tcolorbox}

\usepackage{natbib}  % DO NOT CHANGE THIS AND DO NOT ADD ANY OPTIONS TO IT
\usepackage{caption} % DO NOT CHANGE THIS AND DO NOT ADD ANY OPTIONS TO IT
\usepackage{algorithm}
\usepackage{algorithmic}

\usepackage{newfloat}
\usepackage{listings}
\DeclareCaptionStyle{ruled}{labelfont=normalfont,labelsep=colon,strut=off} % DO NOT CHANGE THIS
\floatstyle{ruled}
\newfloat{listing}{tb}{lst}{}
\floatname{listing}{Listing}
\title{AI-Driven Review Systems:\\ Evaluating LLMs in Scalable and Bias-Aware Academic Reviews}
\author{
    Keith Tyser\textsuperscript{\rm 1}, Ben Segev\textsuperscript{\rm 3}, Gaston Longhitano\textsuperscript{\rm 1}, Xin-Yu Zhang\textsuperscript{\rm 1}, Zachary Meeks\textsuperscript{\rm 1}, Jason Lee\textsuperscript{\rm 1}, Uday Garg\textsuperscript{\rm 1}, Nicholas Belsten\textsuperscript{\rm 5}, Avi Shporer\textsuperscript{\rm 5}, Madeleine Udell\textsuperscript{\rm 4}, Dov Te'eni\textsuperscript{\rm 2}, Iddo Drori\textsuperscript{\rm 1}
}
\affiliations{
    \textsuperscript{\rm 1}Boston University,
    \textsuperscript{\rm 2}Tel Aviv University,
    \textsuperscript{\rm 3}Columbia University,
    \textsuperscript{\rm 4}Stanford University,
    \textsuperscript{\rm 5}MIT\\
}

\begin{document}

\maketitle

\begin{abstract}
Automatic reviewing helps handle a large volume of papers, provides early feedback and quality control, reduces bias, and allows the analysis of trends. Paper reviews are used by researchers and academics, students, lecturers, innovators and entrepreneurs, policymakers and funding agencies, science journalists, and the general public to navigate research, analyze trends, find educational purposes, and find collaborators. We evaluate the alignment of automatic paper reviews with human reviews using an arena of human preferences by pairwise comparisons. Gathering human preference may be time-consuming; therefore, we also use an LLM to automatically evaluate reviews to increase sample efficiency while reducing bias. In addition to evaluating human and LLM preferences among LLM reviews, we fine-tune an LLM to predict human preferences, predicting which reviews humans will prefer in a head-to-head battle between LLMs. We artificially introduce errors into papers and analyze the LLM's responses to identify limitations, use adaptive review questions, meta prompting, role-playing, integrate visual and textual analysis, use venue-specific reviewing materials, and predict human preferences, improving upon the limitations of the traditional review processes. We make the reviews of publicly available arXiv and open-access Nature journal papers available online, along with a free service which helps authors review and revise their research papers and improve their quality. This work develops proof-of-concept LLM reviewing systems that quickly deliver consistent, high-quality reviews and evaluate their quality. We mitigate the risks of misuse, inflated review scores, overconfident ratings, and skewed score distributions by augmenting the LLM with multiple documents, including the review form, reviewer guide, code of ethics and conduct, area chair guidelines, and previous year statistics, by finding which errors and shortcomings of the paper may be detected by automated reviews, and evaluating pairwise reviewer preferences. This work identifies and addresses the limitations of using LLMs as reviewers and evaluators and enhances the quality of the reviewing process.
\end{abstract}


\section{Introduction}
\label{sec:introduction}

The academic community acknowledges the acute need for having foundation models assist reviewing of papers at scale \cite{liu2023reviewergpt,robertson2023gpt4,petrescu2022evolving,schulz2022future,checco2021ai,bao2021predicting,vesper2018peer,latona2024ai,kuznetsov2024can}, along with the risks involved \cite{kaddour2023challenges,spitale2023ai,zou2023universal}. Previous work addresses the limitations of LLM's ability to perform reviewing \cite{liu2023reviewergpt} and their capabilities to review academic papers \cite{liang2023gpt4review}. Large language models demonstrate surprising creative capabilities in text \cite{koivisto2023best}, though they may hallucinate \cite{zhang2023siren}, and demonstrate the power to persuade humans even when inaccurate \cite{spitale2023ai}. This makes controlling the quality and appropriateness of LLM-augmented reviewing highly challenging. At least 15.8\% of reviews for ICLR 2024 were written with AI assistance \cite{latona2024ai}. Recently, an attempt has been made to automate the entire scientific endeavor including generating research ideas, writing code, running experiments, visualizing results, writing scientific papers, and reviewing \cite{lu2024aiscientistfullyautomated}.

Meta-prompting \cite{suzgun2024meta} uses multiple LLM instances for managing and integrating multiple independent LLM queries. Utilizing meta-prompting, the LLM breaks down complex tasks into smaller subtasks handled by expert instances with tailored instructions, significantly enhancing performance across various tasks. This approach outperforms conventional prompting methods across multiple tasks, enhancing LLM functionality without requiring task-specific instructions. Multi-agent review generation for scientific papers \cite{d2024marg} improves LLM reviewing by using multiple instances of LLMs, providing more specific and helpful feedback by distributing the text across specialized agents that simulate an internal discussion. This reduces generic feedback and increases the generation of good comments. Recent work formulates the peer-review process as a multi-turn dialogue between the different roles of authors, reviewers, and decision-makers \cite{tan2024peerreviewmultiturnlongcontext}, and finds that both reviews \cite{latona2024ai} and meta-reviews written by LLMs \cite{santu2024promptingllmscomposemetareview} are preferred by humans over human reviews and meta-reviews.

Why do the Artificial Intelligence, Machine Learning, and Computer Vision communities need AI-based reviews of papers?
(i) AI-based reviews provide early feedback to authors for their work in progress, allowing authors to learn and improve their work;
(ii) AI-based reviews would help conferences maintain high-quality and timely reviews for the increasing number of papers in these fields, as shown in Figure \ref{fig:submissions} in Appendix \ref{appendix:submissions}.
(iii) For quality control of reviews generated while keeping all factors equal;
(iv) So that the entire community can access thousands of reviews for trend analysis and greater paper contextualization.
(v) To reduce human biases in the review process; and 
(vi) To direct readership to high-quality papers based on (AI-based) reviewed merit among the hundreds of thousands of papers available online (for example, the number of papers posted on arXiv grew by 12.3\% from 185,692 in 2022 to 208,493 in 2023).

Paper reviews are used for navigating research, analyzing trends, finding collaborators, adequate citation, and educational purposes. We provide the value of reviews made available to a larger audience using OpenReviewer as an AI reviewing tool to power Papers with Reviews. In particular, we envision the following use cases:
\begin{itemize}
    \item For authors to improve their papers: adequately citing related work, clarity, soundness etc.
    \item For reviewers to help find and refine review points of papers assigned to a reviewer. We note that the recent CVPR 2024 conference banned usage of any LLM for paper reviewing by reviewers, including usage of open-weights LLMs running locally.
    \item To assist conference program chairs or journal editors to quickly identify low-quality works for a desk rejection with human oversight.
    \item For the academic community, a large-scale corpus of reviews of papers on arXiv, delivering free, high-quality reviews based on merit without direct human biases. Currently, arXiv has a total of over 2.5 million submissions, with over 21 thousand papers and over 60 million downloads a month. The academic community selects papers to read based on factors including their field of interest, community discussion, and popularity on social media. Our AI-generated review scores are a valuable metric for selecting which papers to read based on merit rather than popularity and advertising. We review arXiv papers and make the reviews and scores publicly available online.
\end{itemize}

Our key contributions are:
\begin{enumerate}
\item Three AI review systems: (i) OpenReviewer for automatic peer review with LLMs; (ii) Papers with Reviews an online paper review platform; and (iii) Reviewer Arena for evaluating reviewers by preferences.
\item Four evaluation methods: (i) Human evaluation; (ii) Automatic LLM evaluation; (iii) Automatic LLM prediction of human preferences; and (iv) Automatic discovery of LLM review limitations, using synthesis and analysis to map errors and shortcomings in LLM based reviewing.
\item Role playing: Dialogue between LLMs playing different roles in the review process.
\item User feedback: Evaluating quality and trustworthiness.
\end{enumerate}
The paper is structured as follows: Section \ref{sec:review_systems} describes our three review systems. Subsequently, Section \ref{sec:review_evals} describes our four methods for evaluating reviews. Section \ref{sec:review_generation_methods} describes the methods used in generating reviews. In Section \ref{sec:user_feedback_limitations}, we analyze user feedback and address the limitations of our work. Finally, we conclude with our findings and their implications. The supplementary materials consist of 20 Appendices including dataset details, user interface and feedback, prompts, example papers, review questions and scores, evaluation results, and code.

\section{Review Systems}
\label{sec:review_systems}
We present OpenReviewer \footnote{www.openreviewer.com}, Papers with Reviews \footnote{www.paperswithreviews.com}, and Reviewer Arena \footnote{www.reviewerarena.com}.

\subsection{OpenReviewer}

OpenReviewer is a platform designed to augment the traditional peer review process by using LLMs to review papers. Researchers and authors may upload their work to instantly receive peer-review feedback at any point in the writing process. It aims to address the inherent delays and variability in quality of human reviews by providing instant, consistent, high-quality early reviews of academic papers.

We identify requirements for system acceptance by authors and the community, such as journal editors and conference chairs. One requirement is a human-level quality of the review; the other is adherence to community norms of reviewing, e.g., ethical and unbiased reviewing. Human-level reviews are required to make the system trustworthy and informative to the author wishing to revise the paper according to the review. To achieve human-level quality, we design OpenReviewer to consider text and figures in the papers and to adapt the review questions to the journal type and the paper content. To ensure adherence to community norms, we specify appropriate code of ethics and correct for bias in reviewing.

``A picture is worth a thousand words'' expresses succinctly the idea that images convey information more efficiently and persuasively than text. However, when reading scientific papers that include both text and images, it is essential to consider not only pictures or words alone, but the combination of both, which is more powerful than either one alone \cite{schnotz2003construction}. The theory of dual coding \cite{paivio2010dual} explains how humans process images and text so that the combined effect results in better and more efficient comprehension. For instance, image processing is usually fast and holistic, while text (verbal) processing is slow and sequential. For some tasks, but not all, images may be more efficient than text or tables, e.g., detecting a trend. As images and text do not utilize the same processors, readers can process them simultaneously to understand one in the context of the other. Moreover, users often prefer text with pictures over text alone and may regard pictures as more persuasive \cite{powell2015clearer,tal2016blinded}. For these reasons, images are often added to text even though they may not add information already stated in the text.

Previous work on automated reviewing does not consider visual items in papers or map out the errors and shortcomings of papers that LLM reviews can reliably detect. Figures are a significant source of information, and a human reviewer would responsibly review a paper while analyzing its figures. In this work, we consider the paper text and figures within the context of the entire paper while performing an automated review. We introduce errors and shortcomings into papers and evaluate the LLM's ability to detect them by comparing its reviews before and after their introduction, giving a detailed map of which portions of the LLM review are trustworthy.

\begin{figure}[tb]
  \centering
  \includegraphics[width=0.7\linewidth]{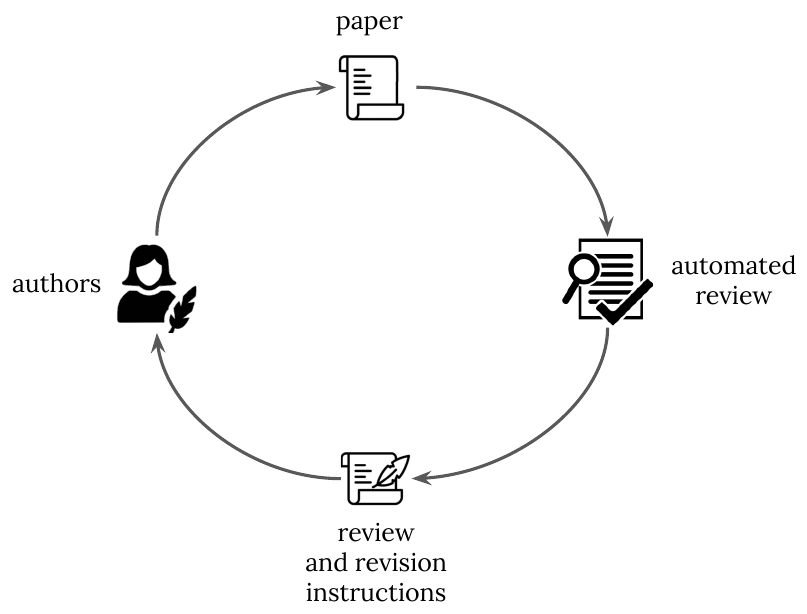}
  \caption{OpenReviewer: A user uploads their paper, which is automatically reviewed, and receives the review along with instructions for revision. The user may provide feedback and upload a revised version. }
  \label{fig:openreviewer}
\end{figure}

\begin{figure}[ht]
\centering
\includegraphics[width=1\linewidth]{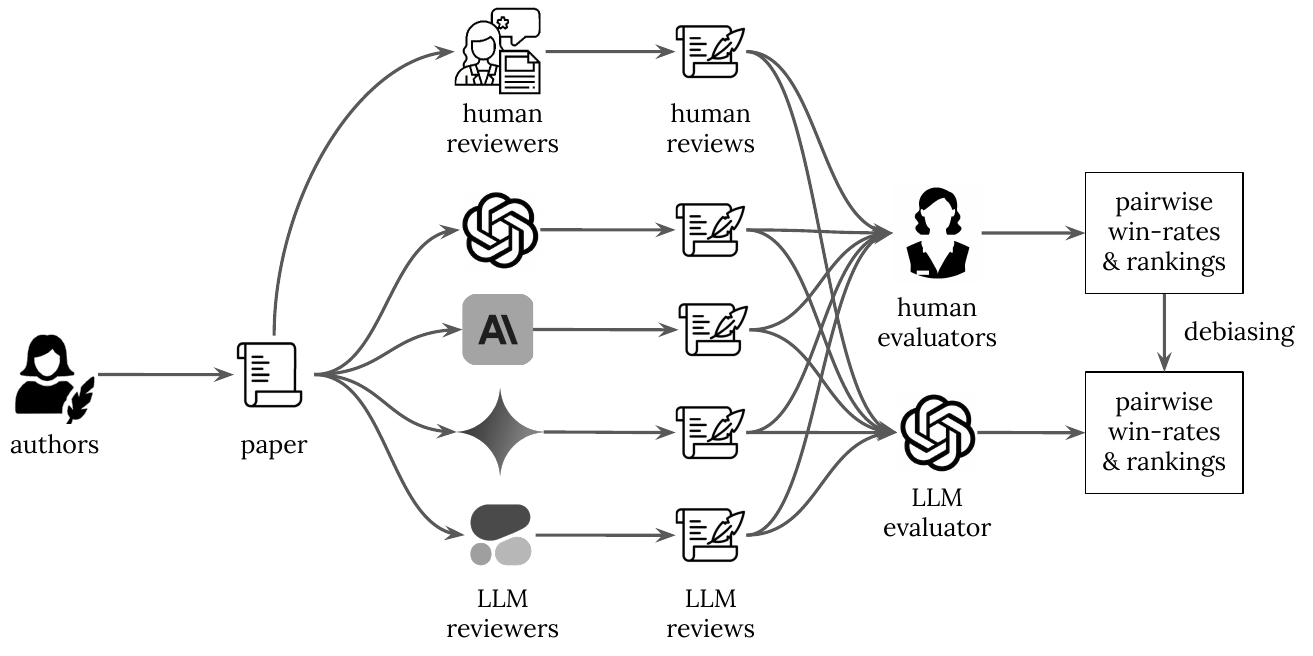}
\caption{Reviewer Arena: The paper is reviewed by human reviewers, three closed LLMs and an open LLM. The reviews are anonymous and human expert evaluators receive pairs of reviews. The experts say whether they prefer one review over another in a Reviewer Arena. The process is repeated using GPT-4 as the expert evaluator. The preferences are used to compute win rate matrices, reviewer scores and rankings.}
\label{fig:reviewerarena}
\end{figure}

We aim to enhance the efficiency and effectiveness of the academic paper review process. We leverage the power of LLMs and user feedback to provide a secure, personalized, and targeted review of academic papers. Our systems, illustrated in Figures \ref{fig:openreviewer} and \ref{fig:reviewerarena}, also allows users to upload their papers. If the user specifies their target conference or venue, then this information enables us to align the review process with the specific guidelines and criteria of the chosen venue. The review generated is then privately provided to the user, giving them constructive feedback to improve their work. Figure \ref{fig:openreviewer} shows a high-level cycle of paper writing, reviewing, and revision, where the system automatically performs the review and provides revision instructions.

\subsection{Papers with Reviews}

\begin{figure}[tb]
  \centering
  \includegraphics[width=1\linewidth]{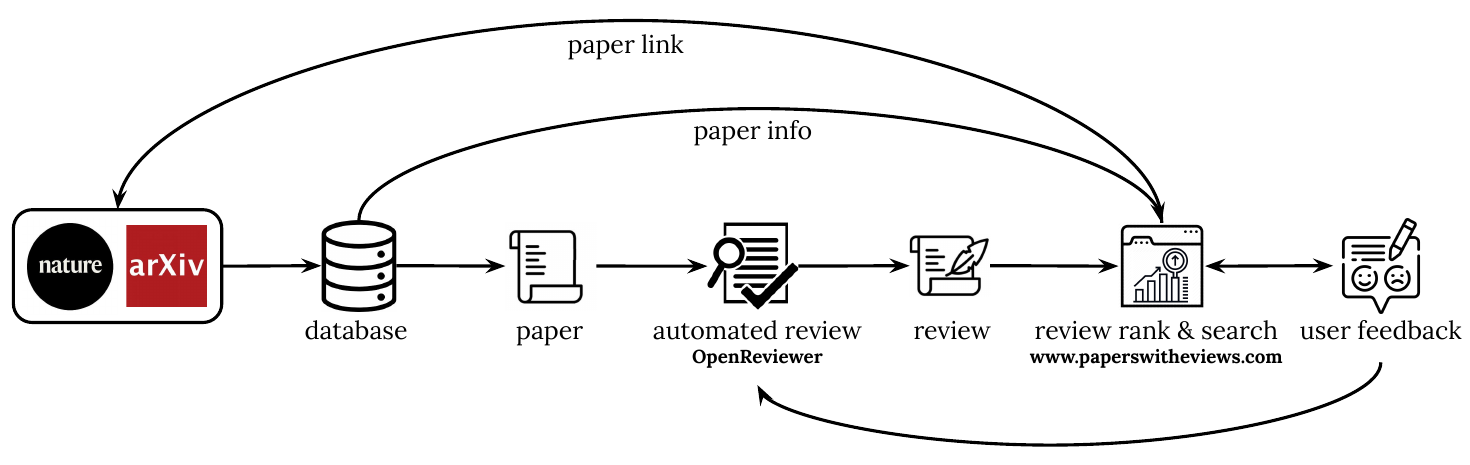}
  \caption{Papers with Reviews: Our system collects papers from arXiv and open-access Nature journals, reviews, ranks, and displays their title, authors, abstract, review, and review score, linking back to the papers on arXiv and Nature. Users provide feedback on the reviews, which is then used to improve the automated review process.}
  \label{fig:archpaperswithreviews}
\end{figure}

We deploy a system called Papers with Reviews illustrated in Figure \ref{fig:archpaperswithreviews} that collects around five hundred academic papers daily from arXiv and around a thousand open-access Nature journal papers monthly. The system reviews and evaluates papers at scale. The system incorporates our LLM-based reviewer, OpenReviewer, that reviews and scores papers. The system has benefits for authors and the scientific community. For authors, the system provides feedback that can be used to improve and revise their work before submission. For the scientific community, the system provides access to high-quality papers of interest based on the paper review scores and enables an analysis of trends and attributes of the growing body of knowledge represented in arXiv and open-access Nature journals. Thus, reviews at scale allow for improved community searching and browsing of papers.

Papers with Reviews complements OpenReviewer by making reviews of publicly available academic papers accessible online. The site targets a broad audience, including researchers, students, and policymakers, offering them valuable insights into the quality and relevance of research papers. By providing LLM-generated reviews alongside human reviews, Papers with Reviews aims to assure the quality and credibility of these evaluations. The service also helps authors improve their work by providing peer-review feedback. It also aids the academic community in navigating the vast landscape of published research, ensuring that high-quality work is recognized and valued appropriately.

Given that effective and trustworthy reviewing at scale is feasible, we demonstrate how Papers with Reviews can benefit the research community. ArXiv Sanity Lite \cite{arxiv-sanity-lite} ranks papers based on publication date, title, abstract, authors, and user tags. More recent papers are given a higher score. User-specified tag strings may be used to rank papers by an SVM classifier trained on TF-IDF vectors derived from paper abstracts, titles, and authors. In addition, papers may be ranked based on the similarity of abstracts and titles between papers. However, arXiv Sanity Lite does not consider or review the paper's content. In contrast, our system, Papers with Reviews, ranks papers based on review scores. These reviews and scores are generated using LLMs based on the entire contents of the paper. Papers with Reviews is currently the only system that publicly provides reviews of thousands of arXiv and open-access Nature papers.

\subsection{Reviewer Arena}

Reviewer Arena is a service for evaluating reviewer quality based on human and LLM preferences, by direct and anonymous comparison of reviews. It is inspired by Chatbot Arena \cite{chiang2024chatbot} which is a platform designed to evaluate LLLMs based on human preferences through pairwise comparisons. Recent progress in human and LLM-generated paper reviews has significantly broadened their applications beyond traditional scholarly assessments, as described above. These developments underscore the potential of leveraging human expertise and LLMs in the task of paper review. However, they also raise important questions about performance evaluation. Traditional benchmarks often need to capture the nuanced insights and varied perspectives that human and machine reviewers bring, especially in assessing their alignment with broader human preferences in diverse academic disciplines. The research community has been exploring innovative benchmarks that evaluate human and LLM reviews to address these challenges. 

This work creates LLM reviews at scale and evaluates the efficacy and alignment of the reviews based on human preferences. Our work introduces a live, dynamic source of review evaluation and utilizes human preferences as a direct metric, aiming to reflect real-world academic review evaluation more accurately. This dynamic approach helps overcome the limitations of static benchmarks, which become outdated and may not effectively capture the interactive and flexible use found in actual review scenarios. By incorporating real-time human feedback into the evaluation process, we can gather a wide range of opinions, enhancing the relevance and applicability of the assessments to real-world academic standards. We introduce a method to facilitate these evaluations, leveraging human and LLM contributions to refine the quality and relevance of academic paper reviews. We coin the term Reviewer Arena for a dynamic and interactive method designed to evaluate and improve the quality of paper reviews through direct anonymous comparison. 
\begin{enumerate}
\item Review Process: We use human and LLM reviews of open conference and journal papers, where they are anonymized to ensure unbiased review. Each paper receives multiple reviews from humans and open and closed LLMs. These reviews cover several dimensions: clarity, originality, significance, and rigor. Human and LLMs reviewers follow conference or journal guidelines.
\item Pairwise Comparison: We employ a pairwise comparison method to evaluate the quality of the reviews. Different reviews of the same paper are presented anonymously to a set of judges: human experts or LLM, who then vote on which review they find more insightful and valuable.
\item Scoring: Based on expert feedback, each reviewer receives a score reflecting its perceived quality and usefulness.
\item Ranking: Reviewers are ranked based on the quality of their reviews. Rankings also serve as a learning tool. Reviewers can see where their reviews stand relative to others and learn from the top-rated reviews.
\end{enumerate}
Our Reviewer Arena aims to enhance the peer review process by making it more transparent, educational, and engaging. This will ultimately raise the overall quality of scholarly feedback and foster a community around constructive critique.

\section{Review Evaluation Methods}
\label{sec:review_evals}
\subsection{Anonymous Human and LLM Evaluation}
To evaluate the quality of the LLM-generated reviews, five expert evaluators were provided with 150 papers together with two anonymous reviews for each paper. Each paper was randomly assigned two reviewers from the list of five potential reviewers: Human, GPT-4 (Turbo-2024-04-09), Claude 3 Opus, Gemini Pro (Bard), and Command R+. The human reviews were obtained from OpenReview submissions. The evaluators were asked which of the two reviews for each paper they preferred. Therefore, this methodology evaluates the relative quality of each reviewer as determined by human evaluators through a series of one-on-one comparisons.

\begin{figure}[htb]
  \centering
  \includegraphics[width=0.9\linewidth]{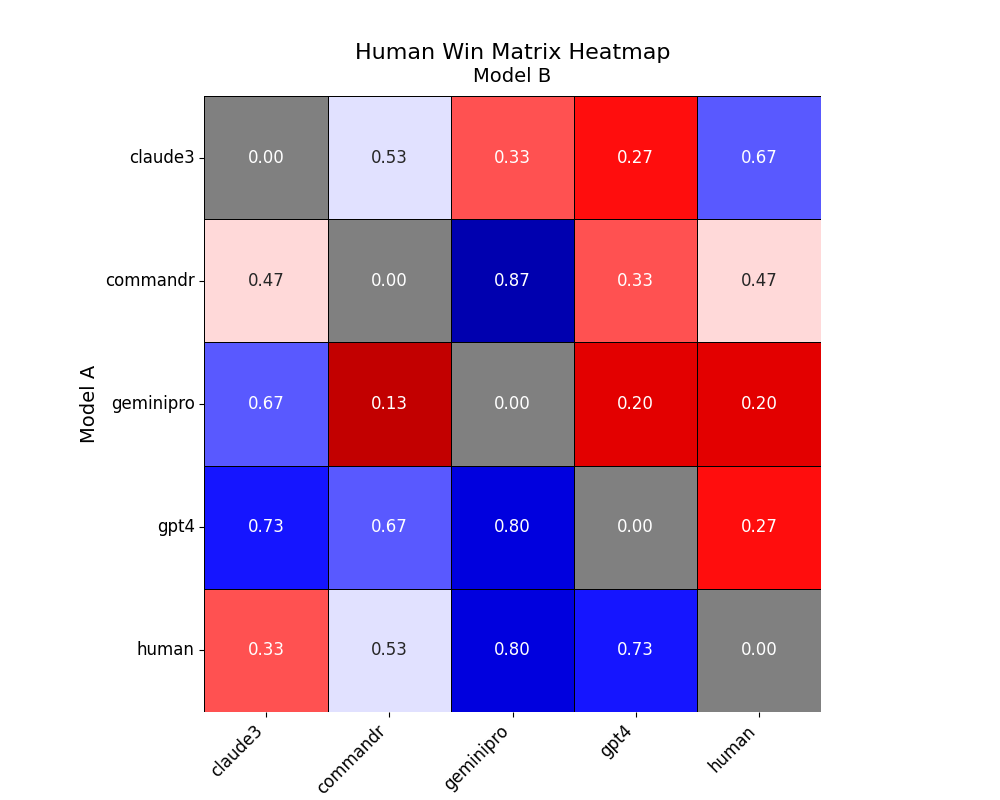}
  \caption{Win rates between five reviewers (three closed LLMs, an open LLM, and a human reviewer) based on human preferences.}
  \label{fig:human-win-matrix}
\end{figure}

This work quantifies and ranks reviewers based on observed match outcomes using a win matrix, Bradley-Terry (BT) model coefficients, and logistic regression. The win matrix represents the outcomes of matches between competitors. For \( N \) competitors, the matrix \( W \) is an \( N \times N \) matrix where each element \( w_{ij} \) represents the probability of competitor \( i \) winning against competitor \( j \), defined as \( w_{ij} = \frac{\text{\# wins of } i \text{ against } j}{\text{total matches between } i \text{ and } j} \). This matrix is constructed by iterating over a list of match results, updating both the win count and the total match count for each pair of competitors. The resultant win matrix for our experiment is show in Figure \ref{fig:human-win-matrix}.

\begin{table}[htb]
\caption{Human preference ranking of reviewers.}
\small
\centering
\begin{tabular}{llc}
\toprule
Rank & Reviewer & Score \\ \midrule
1 & GPT-4 Turbo (April 9, 2024) & 0.558 \\
2 & Human& 0.501 \\
3 & Command R+ & 0.277 \\
4 & Claude 3 Opus & 0.000 \\
5 & Gemini Pro (Bard) & -0.522 \\ \bottomrule
\end{tabular}
\label{tab:human-ranking}
\end{table}

The Bradley-Terry model provides a parametric approach to estimating the relative strengths of competitors based on pairwise comparisons. The probability \( P \) that competitor \( m \) beats competitor \( m' \) follows a logistic function, where $ P\left(H  = \frac{1}{1 + e^{\xi_{m'} - \xi_m}}\right) $
where \( \xi \) represents the vector of BT coefficients, with the constraint \( \xi_1 = 0 \). The BT coefficients are estimated by minimizing the binary cross-entropy loss across all observed matches, where the loss is $\ell(h, p) = -(h \log(p) + (1-h) \log(1-p))$.
The optimization problem is formulated as $\hat{\xi} = \text{argmin}_{\xi} \sum_{t=1}^{T} \ell\left(H_t, \frac{1}{1 + e^{\xi_{A_2} - \xi_{A_1}}}\right)$. We constrain optimization to ensure \( \xi_1 = 0 \). Once the BT coefficients \( \xi \) are estimated, competitors are ranked from strongest to weakest based on these coefficients. The competitors are ranked by sorting the \( \xi \) values in descending order, expressed as \( \text{Ranked Competitors} = \text{sort}(\xi, \text{descending}) \) (Appendix \ref{appendix:code-for-reviewerarena}). The resulting Bradley Terry coefficients and ranking of reviewer for our results are shown in Table \ref{tab:human-ranking}.

\begin{figure}[htb]
  \centering
  \includegraphics[width=0.9\linewidth]{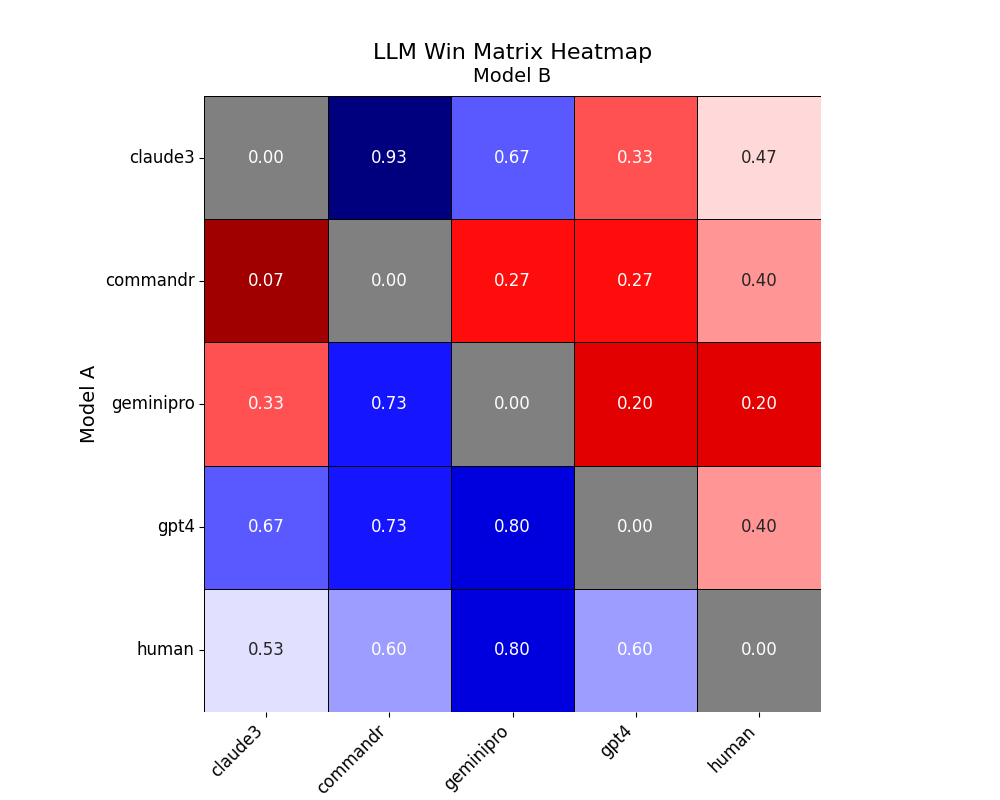}
  \caption{Win rates between five reviewers (three closed LLMs, an open LLM, and a human reviewer) based on GPT 4 Turbo preferences.}
  \label{fig:llm-win-matrix}
\end{figure}

Autoevaluation \cite{boyeau2024autoeval} uses LLM data to evaluate machine learning models, reducing reliance on human-labeled validation data. It is cost-effective and increases the effective sample size, allowing for up to a 50\% increase in efficiency in experiments with GPT-4. The method is based on prediction-powered inference (PPI), which uses limited human data to measure the bias of synthetic data and corrects for it to ensure statistically sound evaluations. Using PPI significantly reduces the variance of estimates and improves the precision of model evaluations without sacrificing accuracy. Inspired by autoevaluation we repeat our evaluation using GPT-4 for providing preferences. Figure \ref{fig:llm-win-matrix} and in Table \ref{tab:llm-ranking}. Using GPT-4 instead of a human evaluator to choose between two reviews allows to use the PPI++ estimate \cite{angelopoulos2023ppi++} of the BT coefficients.

\begin{table}[htb]
\caption{LLM preference ranking of reviewers.}
\small
\centering
\begin{tabular}{llc}
\toprule
Rank & Reviewer & Score \\ \midrule
1 & GPT-4 Turbo (April 9, 2024) & 0.179 \\
2 & Human & 0.119 \\
3 & Claude 3 Opus & 0.000 \\
4 & Gemini Pro (Bard) & -0.819 \\
5 & Command R+ & -1.267 \\ \bottomrule
\end{tabular}
\label{tab:llm-ranking}
\end{table}

\subsection{Automatic LLM Prediction of Human Preference}
We participated in the ``LMSYS - Chatbot Arena Human Preference Predictions'' \cite{lmsys-chatbot-arena} competition, which challenges participants to predict which responses users will prefer in head-to-head comparisons between LLMs. The competition uses a dataset of conversations from Chatbot Arena, where users interact with anonymous LLMs and indicate their preferred responses. The primary goal is to improve LLM alignment with human expectations, which is critical for creating AI systems that better resonate with users. Our success in the competition is largely due to the effective fine-tuning of LLMs, as described in detail in the Appendix \ref{appendix:predicting-human-preferences-implementation-details}. Our focus on alignment is important for ensuring the reliability of AI-generated content in the context of academic review systems.

To predict human preferences between two responses generated by different LLMs, we have the model act as an adjudicator in determining which response aligns more closely with human preferences. This approach aligns with recent studies highlighting the potential of using LLMs not just as content generators but also as evaluators of content quality and alignment with human values \cite{zheng2024judging}. LLMs exhibit various biases that can affect their judgment. For instance, models may exhibit position bias, favoring the first option presented, or verbosity bias, where longer, more detailed responses are preferred irrespective of quality \cite{liu2024lost}; \cite{saito2023verbosity}. There is also a concern with self-enhancement bias, where models may favor responses that self-promote or present themselves in a positive light \cite{zheng2024judging}. 
Our fine-tuned model is adapted to address these biases by training on a curated dataset and incorporating a series of alignment techniques. During inference, we penalize the length of responses and negativity within the responses, as detailed in Appendix \ref{appendix:predicting-human-preferences-implementation-details}. We also correct bias by modifying the resulting predicted probabilities post-processing: (i) Length/Verbosity: We penalize a shorter response \cite{park2024disentangling}; (ii) Sentiment: We use a fine-tuned model \cite{Sanh2019DistilBERTAD} to penalize responses with negative sentiment \cite{gao3improved}; and (iii) Negative patterns: We penalize negative patterns such as phrases ``I'm sorry'' or ``I cannot provide'' based on a correlation analysis.

\subsection{Automatic Discovery of LLM Review Limitations}
\begin{figure}[tb]
  \centering
  \includegraphics[width=1\linewidth]{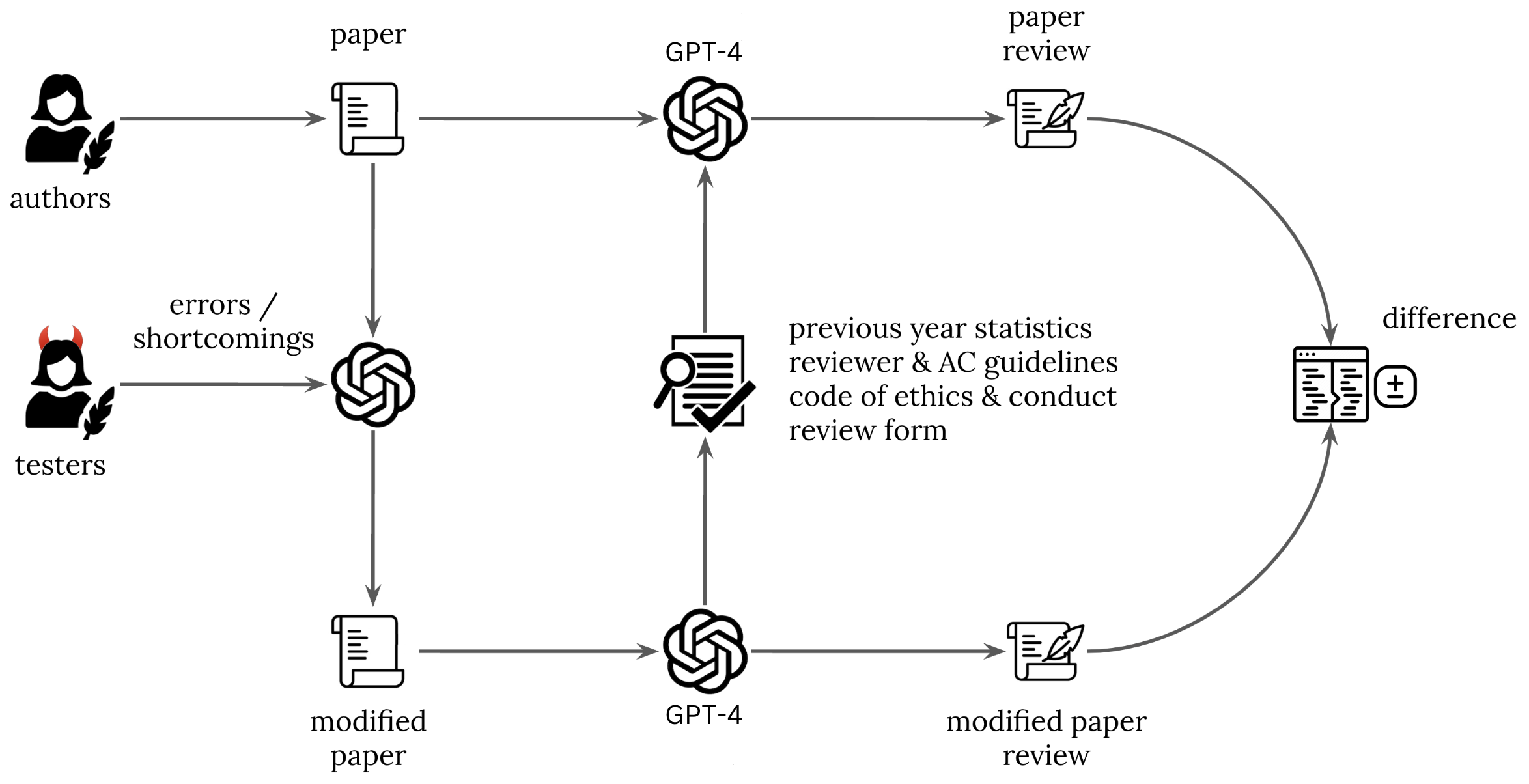}
  \caption{Papers are modified by automatically introducing errors or shortcomings using edit operations, and the LLM reviews the original and modified papers. The original paper review scores are compared with the modified paper review scores, and the content is analyzed to detect the modifications. This process identifies which types of errors and shortcomings the LLM is sensitive to in its review and which types it cannot reliably review, defining the review capabilities and limitations.}
  \label{fig:testers}
\end{figure}

We modify the papers by introducing errors or shortcomings. The LLM reviews the original and modified papers. The original paper reviews are compared with the modified paper reviews by score, and the content is analyzed by detecting the modifications. This process identifies which types of errors and shortcomings the LLM is sensitive to in its review and which types it cannot reliably review, defining the review limitations (Appendix \ref{appendix:analysis-mapping-review-limitations}). Figure \ref{fig:testers} demonstrates how we automatically introduce errors and shortcomings into papers and compare the LLM reviews before and after modification to map the capabilities and limitations of it's reviews. Citations are removed by pattern matching. Theoretical and technical mistakes, metrics, related work, over-claiming, insufficient ablations, lack of comparisons and limitations, are added by prompting the LLM as shown in Table \ref{tab:error-generation} (Appendix \ref{appendix:example-error-and-shortcoming}). Ethical concerns are introduced manually since GPT-4 does not introduce such errors.

\section{Review Generation Methods}
\label{sec:review_generation_methods}
\subsection{Review Questions}
We explored four types of review questions: (i) Fixed questions for a conference or journal: for example, ICLR and NeurIPS papers (Appendix \ref{appendix:open-conference-and-nature-reviews}) have fixed review forms with questions; (ii) Fixed questions for a type of paper: for example, sets of questions for survey, empirical, opinion, etc, papers; (iii) Adaptive choice from a bank of questions based on the paper content: Given the paper and 40 review questions, the LLM selects the top 10 review questions for the paper; and (iv) Adaptive generation of the questions based on the paper content: Given the paper, the LLM generates the top 10 review questions. We applied the fixed questions for maintaining an equal footing and scale. The Appendix \ref{appendix:review-questions} provide details about the review questions.

\subsection{Automatic Summary Comparisons}
\label{sec:automatic-comparison}
\begin{figure}[tb]
  \centering
  \includegraphics[width=1\linewidth]{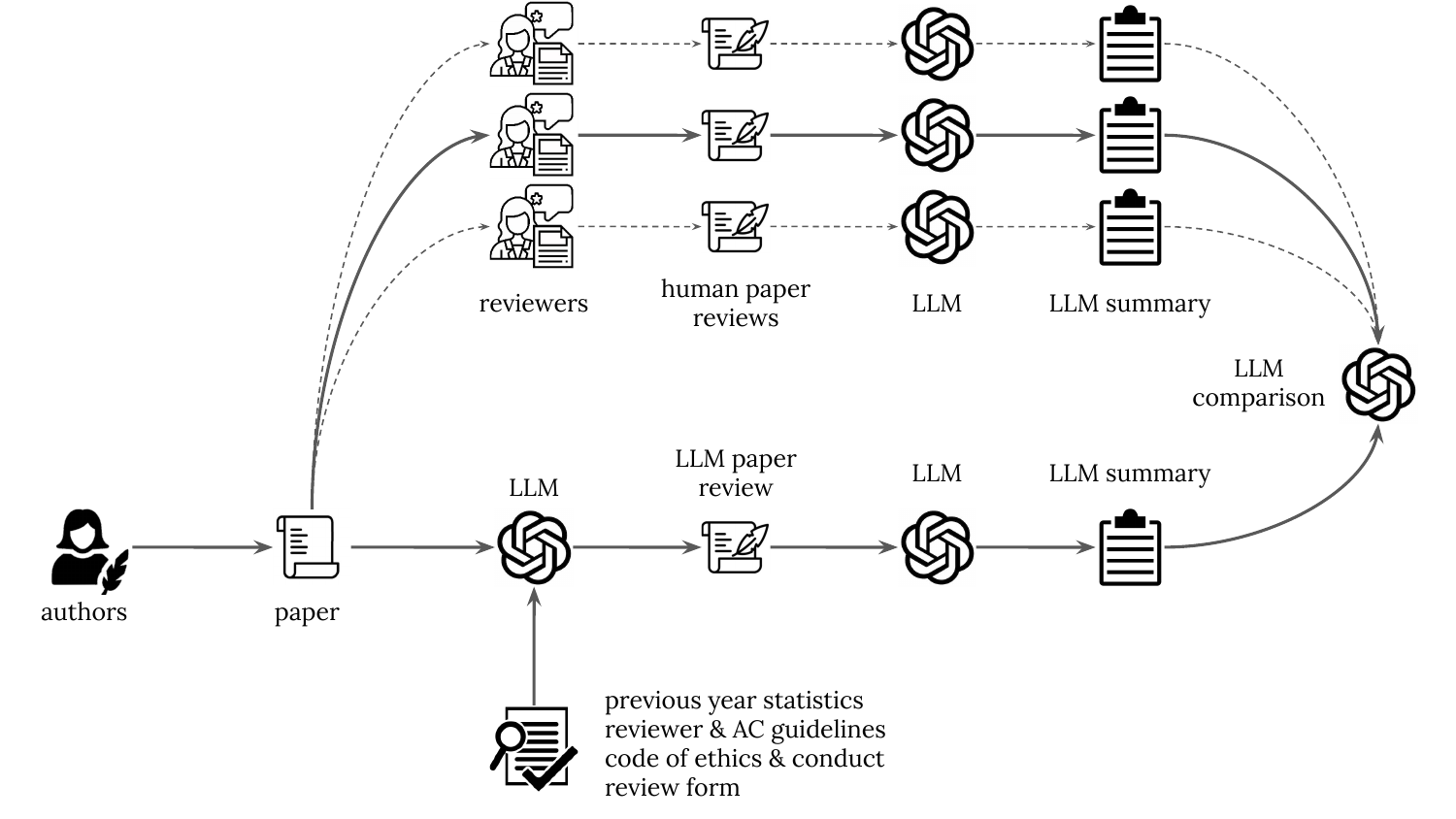}
  \caption{Human reviewers and LLMs review papers. GPT-4 generates a summary for each review by extracting the main points from the reviews. The summary points are compared with one another to find overlap among the human reviews and between the LLM and human reviews.}
  \label{fig:comparisons}
\end{figure}

While LLMs may hallucinate when generating text, they excel at summarizing text as points and comparing between them. We use GPT-4 to extract points of summaries of human and LLM reviews. We use GPT-4 function calling to get structured responses of points and avoid spurious text. Next, GPT-4 compares sets of summary points: human-human and human-LLM reviews. Finally, GPT-4 finds overlapping points and assigns them a similarity score based on how close they are (0-no similarity, 5-identical), providing a measure for comparison (Appendix \ref{appendix:automatic-comparisons}). Figure \ref{fig:comparisons} illustrates the automatic comparison of paper reviews. In a similar fashion, this is useful for automatically comparing between LLM reviews, when performing ablation studies: changing a single factor, such as type and source of data used for generating review, while keeping all other factors equal.

\subsection{Role Playing: Dialogue Between LLMs}
In addition to fixed reviews and adaptive review questions, we experiment with using meta-prompting for playing the roles involved in the human editorial process as shown in Figure \ref{fig:review-process} and described in Appendix \ref{appendix:editorial-review-process}. The process begins with the authors performing science and advertising their work in a paper. A conference has an area chair (AC), and expert reviewers (R), typically at least three, for reviewing the paper. The experts review the paper and submit their reviews. The authors write a rebuttal for each review. Next, the reviewers read the rebuttals and may update their reviews. Based on the reviews, rebuttal, and response, the area chair writes a meta-review and provides a recommendation for a decision to accept or reject, along with a confidence score of this recommendation. Finally, the senior area chair and program chair accept or reject the paper and notify the authors. We simulate the human review process by using the GPT-4 as different personas: program chair (PC), senior area chair (SAC), area chair (AC), reviewers (R), and authors (A). The review simulation consists of multiple steps, roles, and prompts. Figure \ref{fig:llm-editorial-process} in Appendix \ref{appendix:editorial-review-process} illustrates the LLM editorial process, Table \ref{tab:roles} describes the different conference roles and responsibilities, and the simulation is describe by Table \ref{tab:role-playing} in Appendix \ref{appendix:editorial-review-process}.

\begin{figure}[tb]
  \centering
  \includegraphics[width=0.7\linewidth]{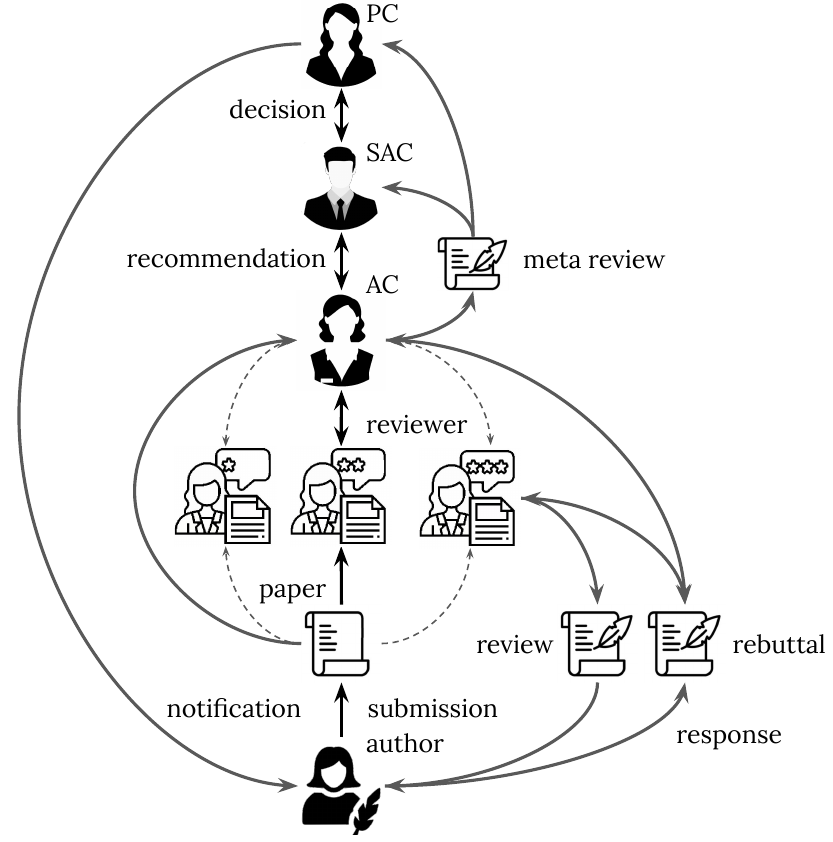}
  \caption{Editorial process: Dialogue between personas - program chair (PC), senior area chair (SAC), area chair (AC), reviewers (R), and authors (A). An LLM simulates each persona. The review process consists of multiple steps: PC-AC assignments, AC-reviewer assignments, reviewing, author rebuttal, reviewer-author discussions, reviewer-AC discussions, meta-reviewing, SAC-AC discussions, SAC-PC decision, and author notification.}
  \label{fig:review-process}
\end{figure}

\section{User Feedback and Limitations}
\label{sec:user_feedback_limitations}
User feedback is used to assess the quality and trustworthiness of the automated feedback and to continually improve the system design. We collect feedback from users in Papers with Reviews. The feedback is on the automated review generated by OpenReviewer for specific papers. The feedback consists of five quantitative questions that evaluate paper reviews \cite{goldberg2023peer} and an open-ended question as described with summary statistics in Appendix \ref{appendix:user-feedback}. 

We use multiple documents related to the review as LLM context: the previous year's statistics, reviewer and area chair guidelines, code of ethics and code of conduct, and the formal review form. These venue-dependent documents result in our review score distributions being similar to human distributions and yielding quality reviews using the full range of scores; however, they require yearly updates. A problem with applying the prediction of human preferences to reviews is that different people may prefer different reviews. A Kaggle competition over a dataset of human preferences provides a common ground for prediction. Correcting for human bias helps partially mitigate this gap, as personal preferences are driven by human behavioral bias.

Future research will extend our evaluation to examine how authors use and trust LLM reviews. Our analysis of the capabilities of LLMs by classifying and testing various reviewing criteria and types of errors and shortcomings indicates the limits of our current application. These limitations are essential for knowing how to use the application, particularly involving the human reviewer. During this work we devised preventive actions for ethical and transparent use of LLMs in reviewing. Future work will also explore self-evolving LLMs for reviewing that independently learn and improve from experience.

\section{Conclusions}

Our aim is to improve scientific writing, research, and communication by providing fast and reliable in-depth reviews on demand. This work evaluates the limitations and capabilities of GPT-4 to review papers and suggest revisions. LLM-generated reviews align well with human reviewers when evaluated by blind human evaluation and an automatic GPT-4 comparison. We present our LLM reviewer system, OpenReviewer, and the associated Papers with the Reviews site. To our knowledge, we are the first to report on such a large-scale empirical evaluation of LLM reviewing.
 
Using human reviews as a baseline, we evaluated value alignment and the process alignment of LLM reviews, i.e., we compared the quality of reviews and the adherence of the reviewing process to conference guidelines and scientific norms of practice. Prior work on LLM academic capabilities suggests that LLMs are now ready for specific reviewing tasks and appear to be more effective for some academic domains and less effective for others \cite{checco2021ai,schulz2022future,liu2023reviewergpt,lu2024aiscientistfullyautomated}. Therefore, we conducted ablation studies and determined the types of errors and shortcomings the LLM can detect and review. When supplied with information about previous editorial decisions, the LLM aligns well with human reviewers. Furthermore, the LLM performs well in detecting specific errors and shortcomings, such as overclaiming, but not others, such as detecting cases in which the authors needed to follow expected norms. We find that iterative design and large-scale empirical evaluation are essential to calibrate the application of LLMs. 

This work leverages LLMs in the review process, addressing challenges and offering proof-of-concept LLM review tools. We introduce and evaluate systems designed to streamline handling tens of thousands of academic papers, from initial collection to reviewing and evaluation. Our methods offer novel approaches to automating academic reviews, improving upon traditional reviews. Our analysis reveals that the system facilitates a more efficient review process and enhances the accessibility and quality of academic literature for both authors and the broader scholarly community. Using papers from arXiv and open-access Nature, coupled with our methods, shows promise in identifying high-quality papers and emerging research trends. In conclusion, our systems and methods represent the different levels of autonomy in the academic review process, detailed in Appendix \ref{sec:autonomy_levels}, and a step forward in automation improvement. With continued development and community involvement, it holds the potential to transform how academic literature is collected, reviewed, disseminated, and evaluated, making it accessible and valuable to researchers worldwide. We hope that our work paves the way for more efficient, consistent, high-quality reviews, accelerating scientific progress while maintaining responsible conduct of research.

\bibliography{aaai25}

\newpage
\clearpage

\onecolumn

\appendix

\section*{Appendices}

\section{Number of Conference Submissions}
\label{appendix:submissions}
Number of submissions to AAAI grew by 12.3\% from 8,777 in 2023 to 9,862 in 2024, NeurIPS grew by 18.6\% from 10,411 in 2022 to 12,345 in 2023, and the number of submissions to CVPR grew by 12.2\% from 8,161 in 2022 to 9,155 in 2023 and by another 25.8\% to 11,532 in 2024.
\begin{figure}[htb]
  \centering
  \includegraphics[width=0.7\linewidth]{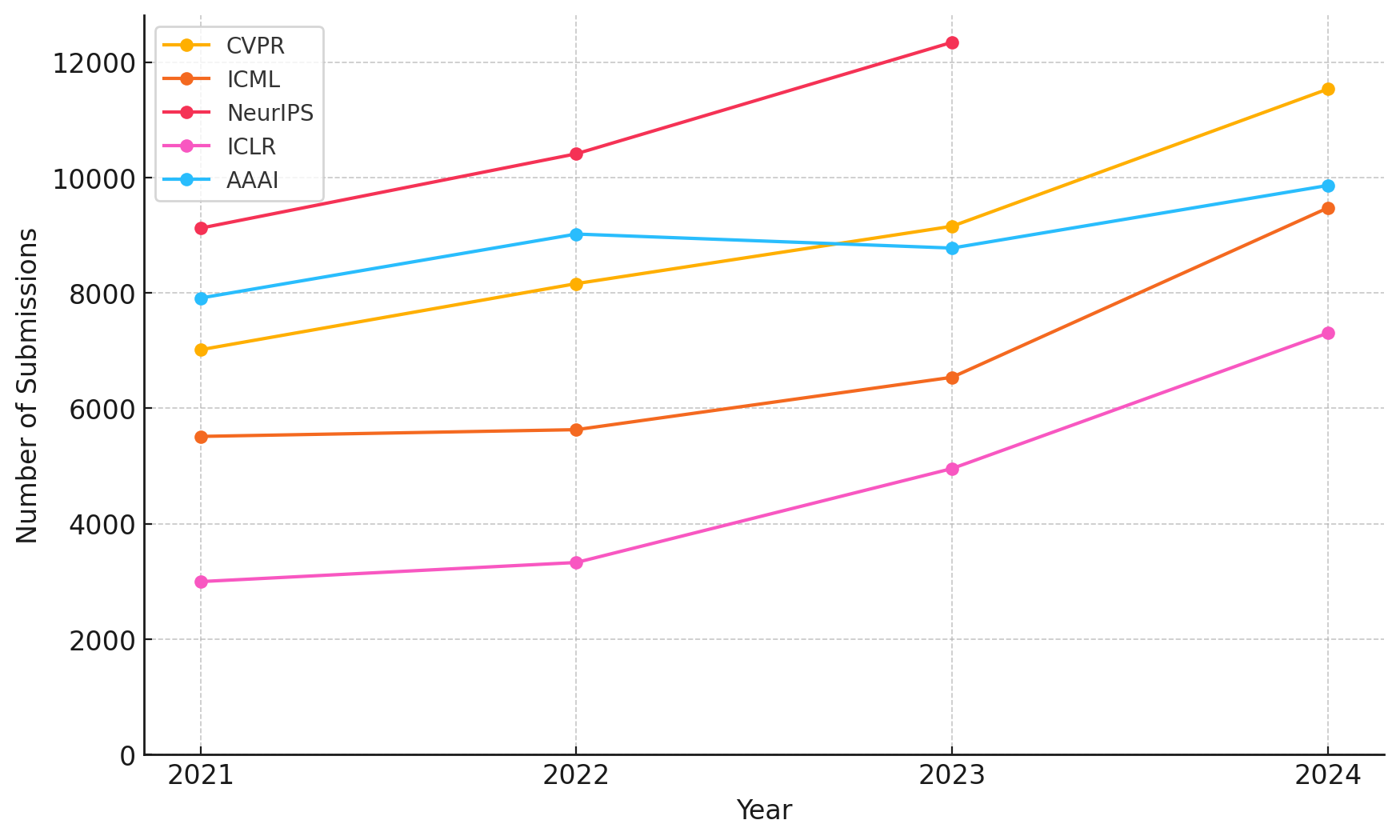}
  \caption{Number of conference submissions by year: CVPR, ICML, NeurIPS, ICLR, amd AAAI.}
  \label{fig:submissions}
\end{figure}

\section{Open-Access Conference and Nature Journals with Reviews}
\label{appendix:open-conference-and-nature-reviews}

\begin{table}[ht]
\caption{Number of papers collected by venue, with open reviews.}
\centering
\begin{tabular}{lc}
\toprule
Source & Number of Papers \\
\midrule
ICLR 2024 & 7404 \\
ICLR 2023 & 4955 \\
NeurIPS 2023 & 12345 \\
NeurIPS 2022 & 10411 \\
\bottomrule
\end{tabular}
\label{tab:papers}
\end{table}

\begin{table}[ht]
    \centering
    \caption{Nature journal IDs and their corresponding names.}
    \begin{tabular}{clcl}
        \toprule
        Journal ID & Journal Name & Journal ID & Journal Name \\
        \midrule
    41467 & Nature Communications & 41594 & Nature Structural \& Molecular Biology \\
    41551 & Nature Biomedical Engineering & 42003 & Communications Biology \\
    41556 & Nature Cell Biology & 42004 & Communications Chemistry \\
    41559 & Nature Ecology \& Evolution & 42005 & Communications Physics \\
    41562 & Nature Human Behaviour & 43246 & Communications Materials \\
    41564 & Nature Microbiology & 43247 & Communications Earth \& Environment \\
    41586 & Nature & 43856 & Communications Medicine \\
    41590 & Nature Immunology &  &  \\
        \bottomrule
    \end{tabular}
    \label{table:journal_names}
    \end{table}

\newpage
\clearpage

\section{User Feedback on Reviews}
\label{appendix:user-feedback}
\begin{figure}[htb]
  \centering
  \includegraphics[width=0.5\linewidth]{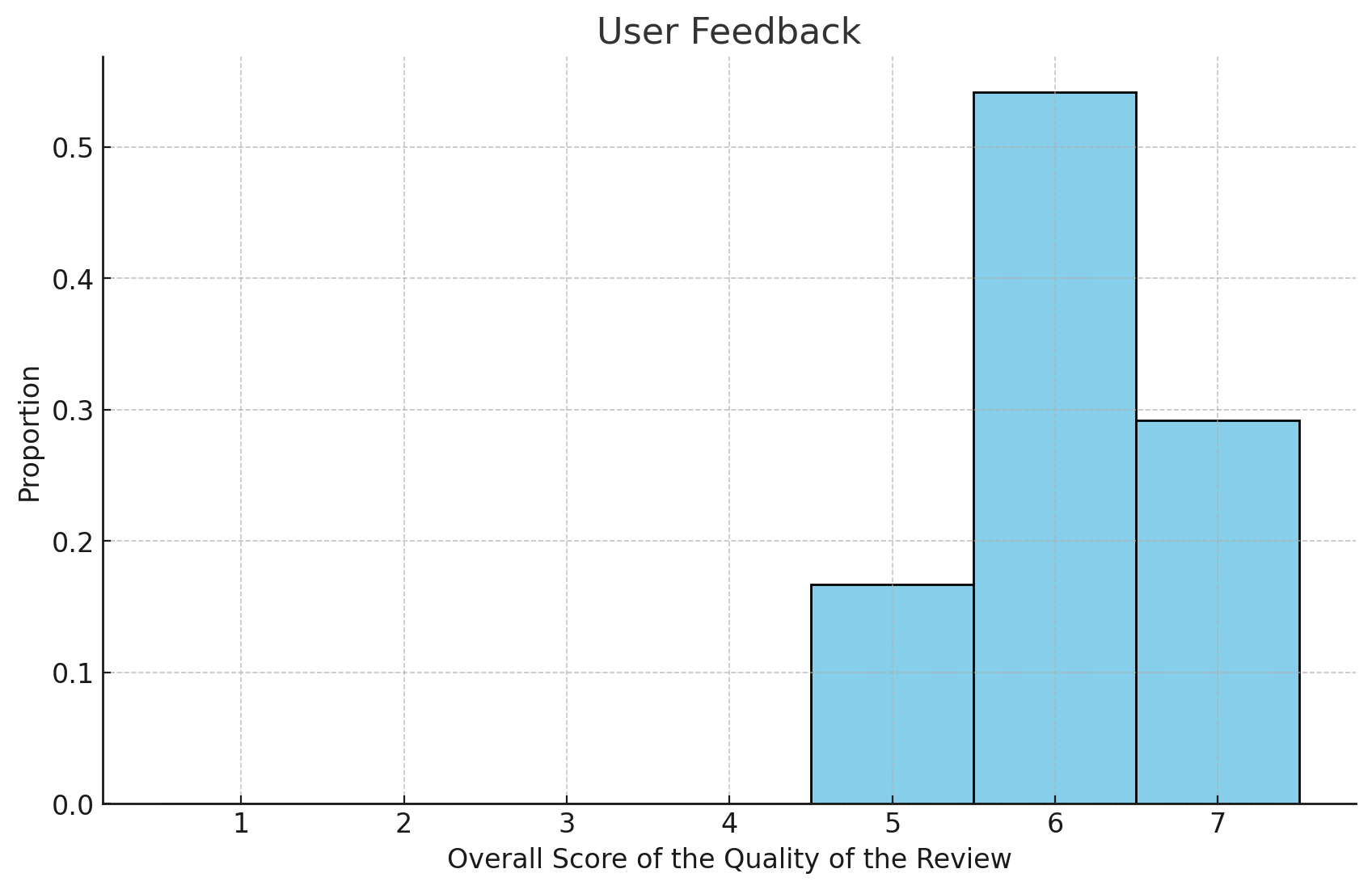}
  \caption{User feedback: Histogram of the scores for the overall quality of the reviews.}
  \label{fig:user-feedback-scores}
\end{figure}

The histogram of the scores for the overall quality of the reviews (not the review scores themselves) is shown in Figure \ref{fig:user-feedback-scores}. The histogram shows the distribution of human feedback scores collected from the site Papers with Reviews. The scores range from 5 to 7, indicating that most reviews are rated very good to exceptional. This demonstrates that the overall quality of reviews is very high, according to the feedback. Summarizing the open-ended feedback provided about the reviews given on Papers with Reviews:
\begin{itemize}
\item Several responses highlight the reviews' detailed, well-organized, and comprehensive nature, noting clear articulation of appreciation, constructive feedback, and identification of drawbacks.
\item The feedback points out the importance of addressing ethical considerations in research, with reviews praised for their emphasis on ethics and suggestions for ethical reviews.
\item A few responses suggest that certain sections, such as correctness, could benefit from more detailed explanations or elaboration.
\item Constructive feedback within the reviews is often recognized for its potential to guide authors in improving their work, including questions that prompt further elaboration on specific aspects of the research or integration into existing frameworks.
\item Specific praise is given for personalized reviews, addressing a paper's unique aspects, such as communication efficiency or the long-term effects of specific methodologies.
\item The feedback also notes the utility of highlighting specific test cases or experimental details from the articles, suggesting this as a strength in understanding and critiquing the material.
\item Feedback suggests that while the reviews help propose additions to the papers, there is a balance to be struck with maintaining focus and conciseness, pointing out that recommendations could be more targeted and specific to enhance their utility.
\end{itemize}
Overall, the feedback reflects an appreciation for the depth and constructiveness of the reviews while also suggesting areas for improvement, such as providing more detailed critiques in certain sections and balancing suggestions for additions with the need for focus and conciseness in the papers.

\begin{figure*}
\begin{tcolorbox}[enhanced, rounded corners,
colback=blue!5!white, colframe=blue!75!black, colbacktitle=blue!85!black,
fonttitle=\bfseries, coltitle=white]
\textbf{Paper Title:} ``Evolutionary roots of the risk of hip fracture in humans'' \\
\textbf{Authors:} Hadas Leah Avni, Nir Shvalb, Ariel Pokhojaev, Samuel Francis, Ruth Pelleg-Kallevag, Victoria Roul, Jean-Jacques Hublin, Frank Rühli, Hila May \\
\textbf{DOI:} s42003-023-04633-4 \\
\textbf{Overall score :} 5 - Very good: A helpful review that stands out on some aspects and provides useful insights. \\
\textbf{Likert scores:} Understanding of the Paper - 4, Coverage of Required Aspects - 4, Support for Evaluations - 3, Constructive Feedback - 3 \\
\textbf{Open feedback on review:} Suggestions for improvement tend to be all additive. Some of the strength of a paper is its focus and conciseness. Recommendations to add content should be balanced by the need to maintain a coherent focus. Recommendations in general are good and grounded in the details of the paper, however they do not often address specific locations or content within the paper. This is one of the most helpful features of peer reviews which is largely missing.
\end{tcolorbox}
\caption{Example user feedback on review.}
\end{figure*}

\begin{figure*}
\begin{tcolorbox}[enhanced, rounded corners,
  colback=blue!5!white, colframe=blue!75!black, colbacktitle=blue!85!black,
  fonttitle=\bfseries, coltitle=white]
\textbf{Paper Title:} ``Uncertainty-Aware Deep Attention Recurrent Neural Network for Heterogeneous Time Series Imputation'' \\
\textbf{Authors:} Linglong Qian, Zina Ibrahim, Richard Dobson \\
\textbf{arXiv ID:} 2401.02258 \\
\textbf{Overall score :} 7 - Exceptional: An excellent review, helps authors to non-trivially improve the paper and brings a unique piece of information that is crucial for the decision. \\
\textbf{Likert scores:} Understanding of the Paper - 5, Coverage of Required Aspects - 5, Support for Evaluations - 5, Constructive Feedback - 5 \\
\textbf{Open feedback on review:} The review seems really detailed and well-defined/ordered to me. Everything is stated very clearly whether it is appreciation or constructive feedback or explaining the drawbacks.
\end{tcolorbox}
\caption{Example user feedback on review.}
\end{figure*}

\begin{figure*}
\begin{tcolorbox}[enhanced, rounded corners,
  colback=blue!5!white, colframe=blue!75!black, colbacktitle=blue!85!black,
  fonttitle=\bfseries, coltitle=white]
\textbf{Paper Title:} ``Coefficient Shape Alignment in Multivariate Functional Regression'' \\
\textbf{Authors:} Shuhao Jiao, Ngai-Hang Chan \\
\textbf{arXiv ID:} 2312.01925 \\
\textbf{Overall score :} 7 - Exceptional: An excellent review, helps authors to non-trivially improve the paper and brings a unique piece of information that is crucial for the decision. \\
\textbf{Likert scores:} Understanding of the Paper - 4, Coverage of Required Aspects - 4, Support for Evaluations - 5, Constructive Feedback - 5 \\
\textbf{Open feedback on review:} The best thing about this review was the feedback given to the authors about mentioning the ethical impacts of their research (mentioned in heading 6 - Correctness).
\end{tcolorbox}
\caption{Example user feedback on review.}
\end{figure*}

\newpage
\clearpage

\section{Review Scores}
\label{appendix:review-scores}
\begin{figure}[htb]
  \centering
  \includegraphics[width=\linewidth]{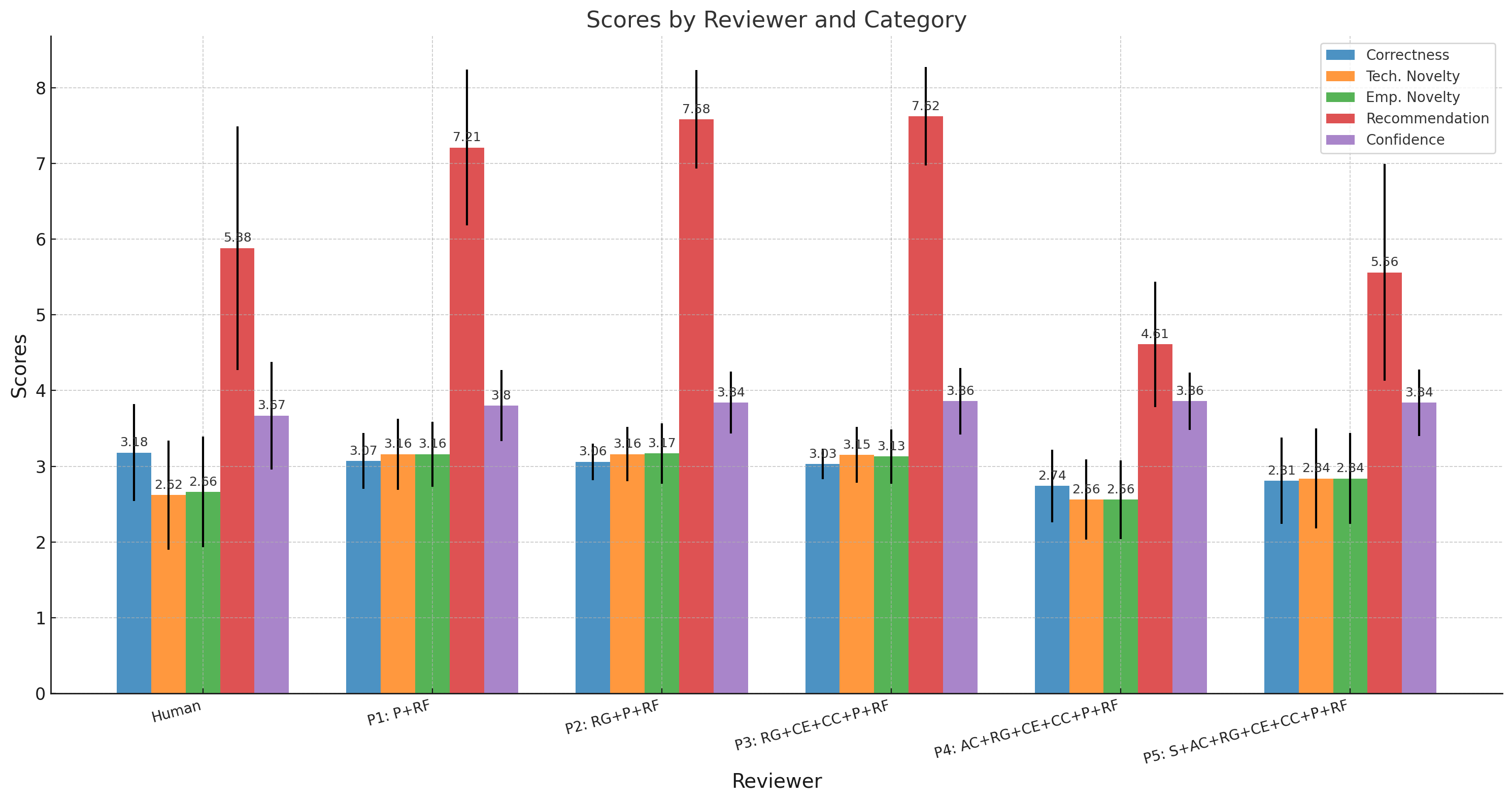}
  \caption{Ablation of in-context LLM review scores: Average and standard deviation scores of the human reviewers and LLM review for paper correctness, technical novelty and significance, empirical novelty and significance, overall recommendation score, and confidence. P1, P2, P3, P4, P5 ablate the increasing documents used in the GPT-4 context prompt. P1 includes the full paper text (P) and conference review form (RF). P2 adds the reviewer guide (RG). P3 adds the code of ethics (CE) and code of conduct (CC). P4 adds guidelines for the area chair (AC). P5 adds the statistics of the previous year's conference.}
  \label{fig:human-p12345}
\end{figure}

Figure \ref{fig:human-p12345} shows the average and standard deviation scores of the human reviewers and LLM review for paper correctness, technical novelty and significance, empirical novelty and significance, overall recommendation score, and confidence. P1, P2, P3, P4, P5 ablate the increasing documents used in the GPT-4 context prompt. P1 includes the full paper text (P) and conference review form (RF). P2 adds the reviewer guide (RG). P3 adds the code of ethics (CE) and code of conduct (CC). P4 adds guidelines for the area chair (AC). P5 adds the statistics of the previous year's conference.

The human reviewers have an average recommendation score of 5.88, with a standard deviation 1.61. With the context of the entire paper text and the conference review form (P1), the LLM has an average recommendation of 7.21, higher than the human reviewers. The standard deviation of 1.03 is less than that of the human reviewers. Adding the reviewer guide to the context (P2) slightly increases the recommendation score to 7.58. The standard deviation is reduced further, with a more consistent scoring by the LLM. With the addition of the code of ethics and code of conduct to the context (P3), the recommendation score slightly increases to 7.62, similar to P2, and the standard deviation remains the same. After adding guidelines for the area chair (P4), the recommendation score decreases to 4.61, indicating that this context makes the LLM more critical or stringent in its evaluations due to the knowledge of expected outcomes. With the addition of the previous year's conference statistics (P5), the recommendation score improves and is near the human reviewer's score. The standard deviation also increases, indicating more variability in the scoring.
In summary, LLM contexts P1, P2, and P3 consistently give higher recommendation scores than the human reviewers, suggesting a more positive or lenient view of the papers. P4 context, with the area chair guidelines added, shows a significant decrease in recommendation scores, suggesting these guidelines influence the LLM to be more critical in its evaluations. P5 reaches the same level of recommendation as the human reviewers.

To examine the reviews further, we compared the score distributions of GPT-4 with all documents (P5) and the human reviewers as shown in Figures \ref{fig:human-p5-distributions-scores} and \ref{fig:human-p5-distributions-rec} of the supplementary material. GPT-4 P5 score distributions were similar to human scores for correctness, technical and empirical novelty, and significance; however, they were skewed to higher values compared with the human distributions for confidence. The overall recommendations of P5 and human reviews have a similar mean and standard deviation.  

\newpage
\clearpage

\section{Evaluating Reviews}
\label{appendix:evaluating-reviews}
The human review evaluator assesses reviews written by human reviewers and the LLM, GPT-4 with context P5. The human review writer is an ICLR 2023 reviewer. Table \ref{tab:humaneval} shows the average evaluation results on a randomized sample of 5\% of the papers evaluated by human experts.

\begin{table*}[htb]
\caption{The human review evaluator evaluates human and P5 written reviews of papers. The human review writer is an ICLR 2023 reviewer. The LLM is GPT-4 with context P5. The evaluation is on a scale of 0-5 (0 being the worst, five the best). For the third question, a score of 0 indicates a content-free review.}
\centering
\begin{tabular}{lcccc}
\toprule
Review Evaluator: & Human & Human & GPT-4 & GPT-4 \\ \midrule
Review Writer: & Human & P5 & Human & P5 \\ \midrule
How well does the review explain the score? & 4.80$\pm$0.39 & 4.76$\pm$0.51 & 4.27$\pm$0.65 & 4.65$\pm$0.52 \\ \midrule
How well does the review guide the authors to improve the paper? & 4.66$\pm$0.51 & 4.79$\pm$0.71 & 4.14$\pm$0.50 & 4.27$\pm$0.45 \\ \midrule
Does the review contain content specific to the paper? & 4.53$\pm$0.79 & 4.68$\pm$0.82 & 4.97$\pm$0.16 & 4.95$\pm$0.22 \\ \bottomrule
\end{tabular}
\label{tab:humaneval}
\end{table*}

\paragraph{Confusion matrix.}
Considering the average human review rating as ground truth, we analyze false positives and negatives. We consider the LLM's two types of errors: accepting a paper rejected by human reviewers and rejecting a paper that human reviewers accepted. One paper that the LLM accepted with a score of at least 7 was rejected by the average of the human reviewers with a score of at most 3. Four papers that the LLM rejected with a score of at most 3 were accepted by the human reviewers with a score of at least 7. Eight papers that the LLM accepted with a score of at least 6 were rejected by the average of the human reviewers with a score of at most 4. 22 papers that the LLM rejected with a score of at most 4 were accepted by the human reviewers with a score of at least 6.

\section{Blind Human Evaluation}
\label{appendix:blind-human-evaluation}
In explaining the review score, reviews written by humans and GPT-4 with all related review contexts were assessed close to each other by human evaluators, with scores of 4.80 and 4.76. Reviews written by humans but evaluated by GPT-4 scored lower at 4.27. Reviews written by GPT-4 and evaluated by itself scored 4.65. In guiding authors to improve their papers, reviews written by GPT with all related review contexts and evaluated by humans scored the highest at 4.79. The lowest score of 4.14 was for reviews written by humans but evaluated by GPT. Regarding content specificity, GPT, when evaluating human-written reviews, provided the highest score of 4.97. Reviews written by GPT and evaluated by humans scored slightly lower at 4.68. A score of 0 in this context would indicate a content-free review; however, all scores are considerably higher than that. There was one instance where GPT accepted a paper with a high score of at least 7, but human reviewers collectively disagreed and gave it a low score of at most 3. Conversely, GPT rejected four papers with low scores (at most 3), which human reviewers found to be of high quality, scoring them at least 7. A slightly less strict threshold showed that GPT accepted eight papers with a score of at least 6 which human reviewers rated poorly (at most 4). On the other hand, 22 papers that LLM gave low scores to (at most 4) were considered of good quality by human reviewers, giving them scores of at least 6.

\newpage
\clearpage

\section{Synthesis and Analysis for Mapping Review Capabilities and Limitations}
\label{appendix:analysis-mapping-review-limitations}
\begin{figure}[htb]
  \centering
  \includegraphics[width=0.9\linewidth]{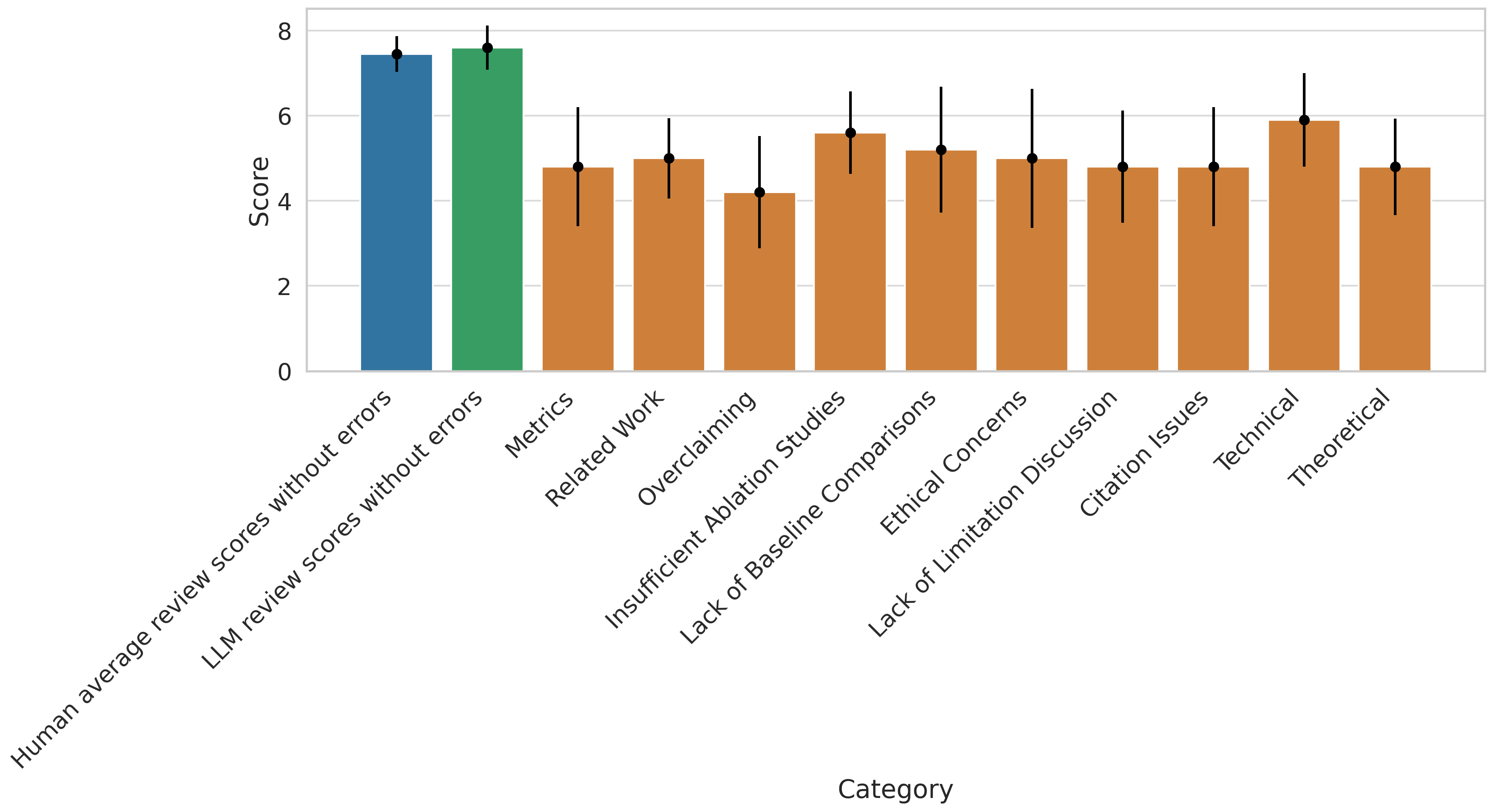}
  \caption{Average human and LLM review scores of the original papers, and average LLM review scores for each type of error or shortcoming introduced into the papers. Lower scores (in orange bars) are better, representing the LLM's ability to detect errors or shortcomings and decrease the review score due to those type of error.}
  \label{fig:errors}
\end{figure}

Figure \ref{fig:errors} shows average review scores for various types of errors and shortcomings introduced into the papers, with error bars showing the standard deviation. The human and LLM average review scores of the original papers without errors and shortcomings are highlighted with distinct colors. The human average review scores without errors and the LLM review scores without errors are close, indicating a general alignment in their evaluations. The overclaiming category has relatively lower average scores, indicating that the LLM review easily detects these errors and reduces the scores.

\begin{figure}[htb]
  \centering
  \includegraphics[width=0.9\linewidth]{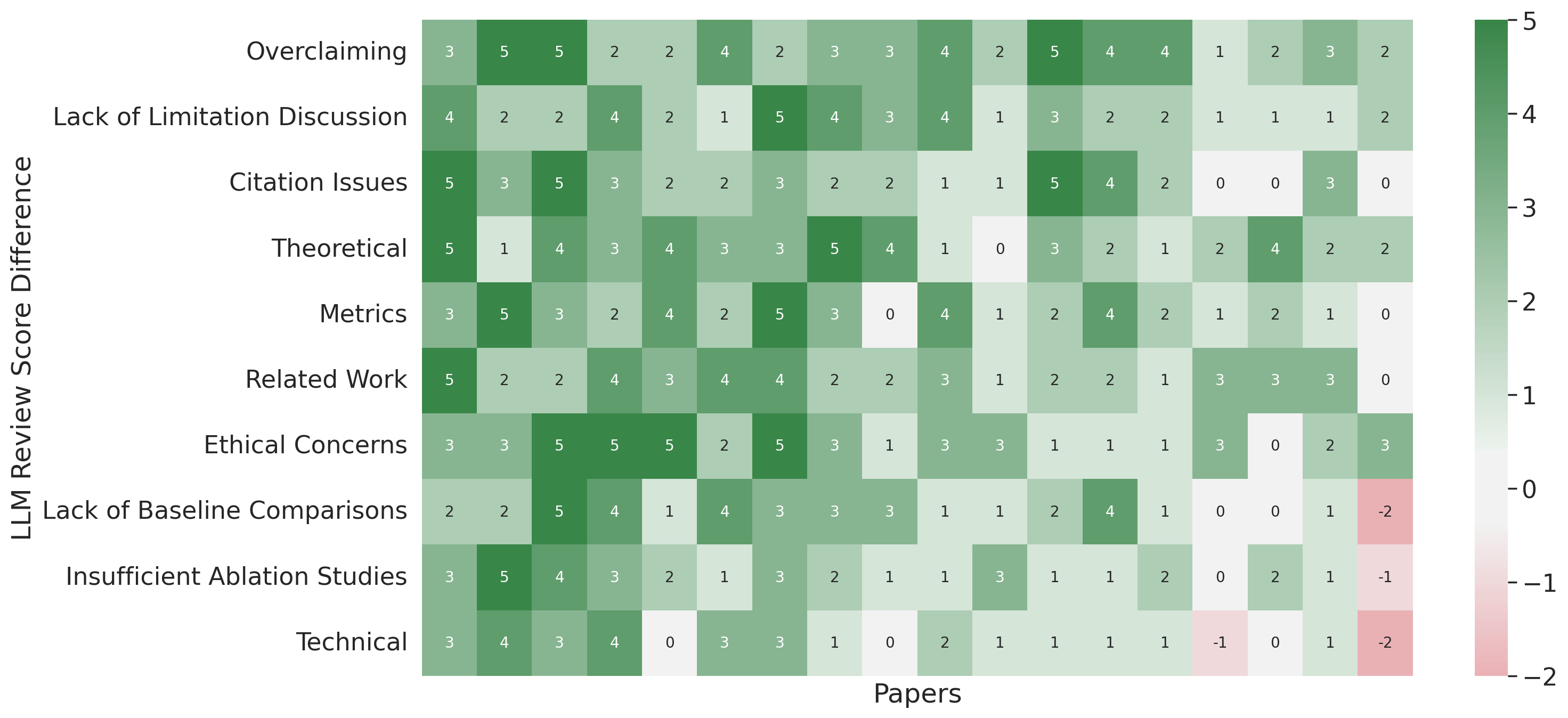}
  \caption{Heatmap of difference between LLM review scores with and without errors, by magnitude over errors categories and papers (green is better). The color gradient, ranging from green to red, indicates how well the LLM detects its limitations by modifying the review score before and after the injection of the errors. }
  \label{fig:heatmap}
\end{figure}

\begin{figure}
  \centering
  \includegraphics[width=0.6\linewidth]{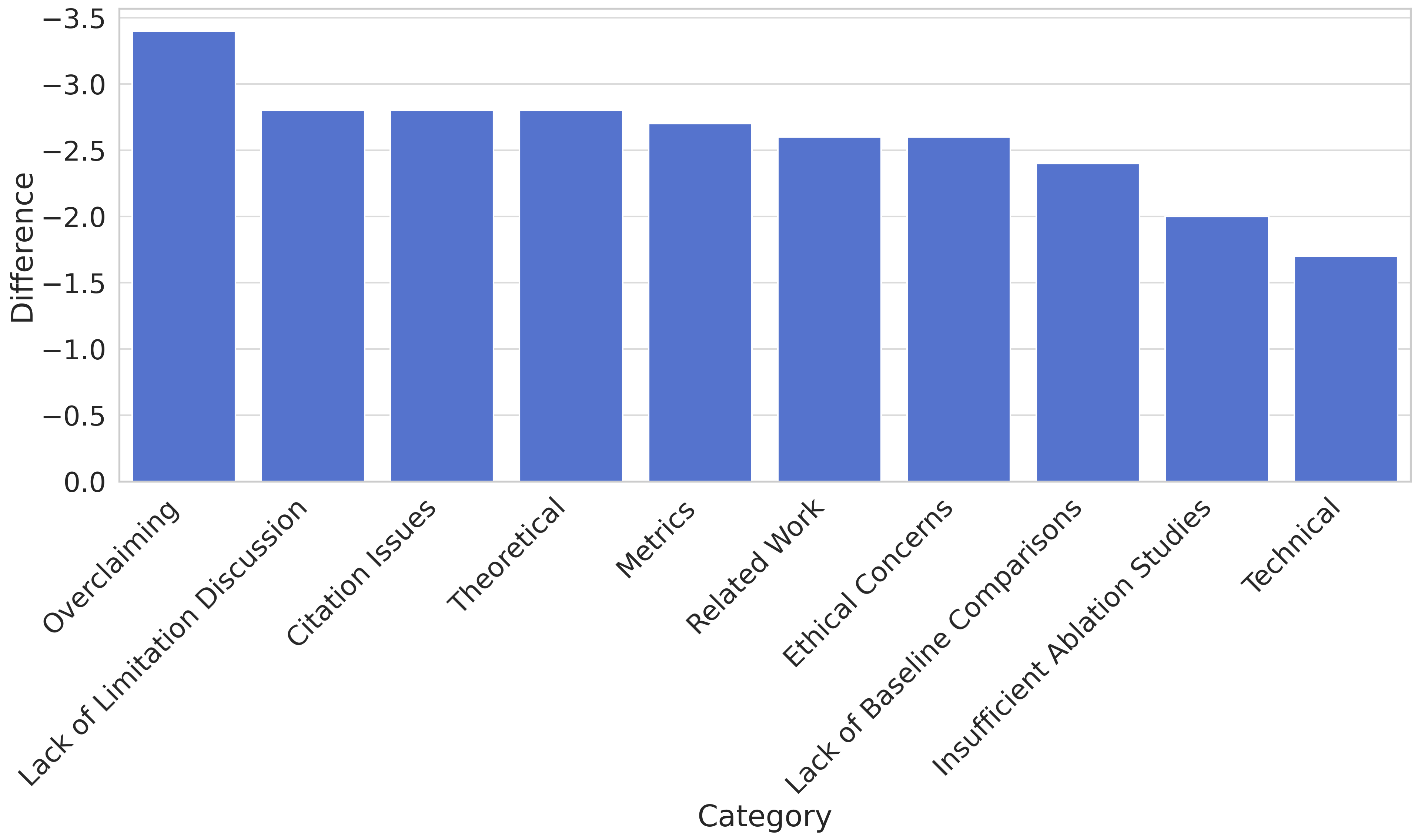}
  \caption{Average LLM review score penalty for different types of errors or shortcomings. The LLM review detects the different types of errors and markedly penalizes overclaiming.}
  \label{fig:diffs}
\end{figure}

Figure \ref{fig:diffs} shows the difference in LLM review scores for various error and shortcoming categories compared to the LLM review scores without errors. Figure \ref{fig:heatmap} shows a heatmap using red and green colors to indicate non-positive and positive difference values. The intensity of the color corresponds to the magnitude of the difference between the LLM review score of the original and modified papers. Most of the data points are positive, indicating correct error detection. The categories with the highest detection are overclaiming across most of the papers. Citation issues and Technical errors also stand out, with several papers having higher values, indicating the detection of these errors. Ethical Concerns and Insufficient Ablation Studies have low values or zeros. The heatmap visually summarizes the trustworthiness of reviews across different categories and papers and categories. The LLM is good at finding Overclaiming, Citation Issues, and Theoretical. In contrast, Ethical Concerns and Technical may be overlooked. The ability of the LLM also varies for different papers. The LLM usually give itself a high confidence score in its rating, therefore knowing these difference values for each paper is essential for understanding which parts of the review can be trusted.

Lack of Baseline Comparisons and Metrics shows significant variability across papers. Technical has mostly low values across papers, suggesting that the technical errors in many paper reviews are not detected. Certain columns representing specific papers have many rows with higher values, indicating that errors were detected across most categories in those particular papers. Some columns have predominantly lower values, suggesting that in the corresponding papers, errors were not detected.

We introduce errors and shortcomings into papers by deleting, inserting, or editing text using GPT-4 by adversarial prompts. Figures \ref{fig:errors-related-work}, \ref{fig:errors-baseline}, \ref{fig:errors-citations}, \ref{fig:errors-technical-original}, \ref{fig:errors-technical-modified}, \ref{fig:errors-overclaim1-orig}, \ref{fig:errors-overclaim1-mod}, \ref{fig:errors-overclaim2-orig}, \ref{fig:errors-overclaim2-mod} show side-by-side examples of an original paper (which received a review score of 7) and edits deleting related work (reducing the score to 5), removing baseline comparisons (reducing the score to 4), removing citations (reducing the score to 5), modifying equations (reducing the score to 5), and over-claiming (reducing the score to 3). The deletions are highlighted in orange and the modifications in blue.

Table \ref{tab:error-generation} describes the classification of different types of errors or shortcomings in papers and how these are introduced into papers. Figure \ref{fig:diffs} shows the deviations of each category from the average LLM review scores of the original papers. The differences are sorted by magnitudes, showing that the LLM review detects the different types of errors, reducing their review scores and specifically penalizes overclaiming.

\newpage
\clearpage

\section{Examples of Error and Shortcomings and Adversarial Prompts}
\label{appendix:example-error-and-shortcoming}
\begin{table*}
\caption{We classify different types of errors in papers and then introduce these errors into papers. OpenReviewer reviews the papers without and with the errors. We compare the reviews of the original papers and the reviews of papers with the errors, and finally, we detect the errors in the text of the review of the papers with the errors and their scores. Theoretical mistakes and technical errors are further classified into sub-types. Eight error and shortcoming types are introduced using GPT-4, ethical errors are introduced manually, and citations are removed by pattern matching.}
%\scriptsize
\centering
\begin{tabular}{p{3.5cm}p{6.5cm}p{6.5cm}}
    \toprule
Error or Shortcoming & Description & Example\\
    \midrule
Theoretical Mistakes & These can range from incorrect mathematical derivations to making unfounded assumptions about a model or algorithm. & LLM/Human: Generate a situation where the paper includes common theoretical errors observed in submission. These errors should include incorrect mathematical derivations, unfounded assumptions, misinterpretations of existing theories, lack of theoretical justification, and ambiguous definitions.
\\
\midrule
Metrics & 
Not reporting important metrics or details about experiments. & Revise the paper by removing metrics in the experiments.\\
\midrule
Related Work & Not comparing with the state-of-the-art or relevant baselines. & Remove the Related Work section from the paper.\\
\midrule
Overclaiming & Making exaggerated claims about the novelty or impact of the work without substantial evidence can be problematic. & Exaggerate the paper's initial assertions or incorporate over-claiming statements into the paper.\\
\midrule
Insufficient\newline Ablation Studies & Ablation studies help demonstrate which components of a proposed system contribute to its performance. Without these, it can be hard to understand the significance of the introduced changes. & Remove ablation studies from the paper.\\
\midrule
Lack of\newline Baseline Comparisons & Not comparing with standard or widely-accepted baseline methods can contribute to seeming less grounded. & Remove comparisons with standard or accepted baseline methods.\\
\midrule
Ethical Concerns & Not addressing potential ethical implications of the work, especially if the work touches on sensitive areas like facial recognition, can be a red flag. & Human\\
\midrule
Lack of Discussion on Limitations & Every approach has limitations. Not discussing them or addressing potential criticisms can show a lack of thoroughness. & Remove text discussing the limitations of the work from the paper.\\
\midrule
Citation Issues & This includes not citing relevant prior work, which can make it seem like the authors are unaware of the literature or over-citing one's previous work without reason. & Pattern\newline matching\\
\midrule
Technical Errors & These can range from coding mistakes in the provided implementation to incorrect use of statistical tests. & Include technical errors, mistakes that range from incorrect use of statistical tests to evaluation errors, and issues with parameter tuning and model selection. \\
  \bottomrule
\end{tabular}
\label{tab:error-generation}
\end{table*}

\begin{figure*}[htb]
\centering
  \includegraphics[width=0.9\linewidth]{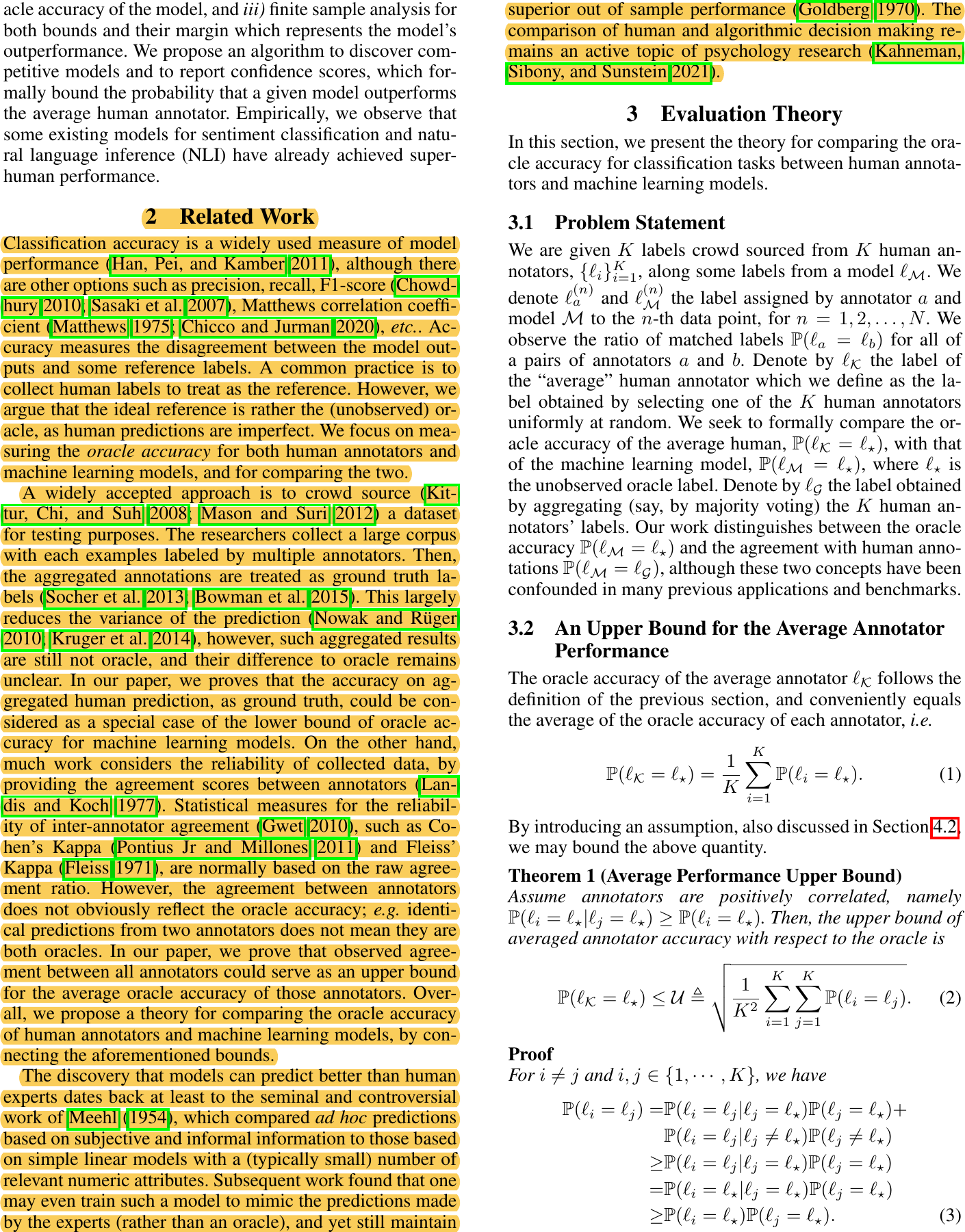}
  \caption{Example of introducing related work errors or shortcomings: Deleting related work.}
  \label{fig:errors-related-work}
\end{figure*}

\begin{figure*}[htb]
\centering
  \includegraphics[width=0.9\linewidth]{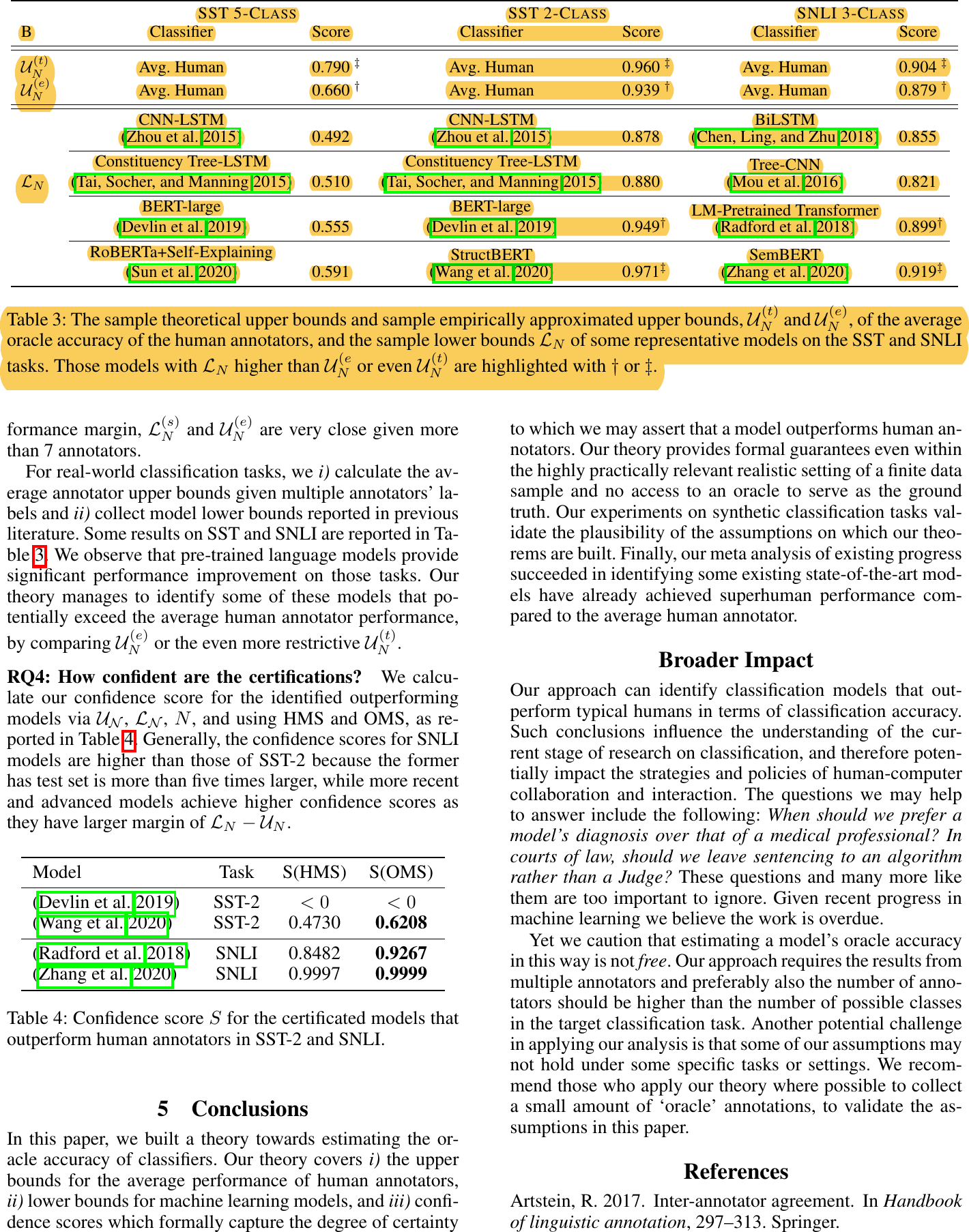}
  \caption{Example of introducing baseline errors or shortcomings: Deleting baseline comparison.}
  \label{fig:errors-baseline}
\end{figure*}

\begin{figure*}[htb]
\centering
  \includegraphics[width=0.9\linewidth]{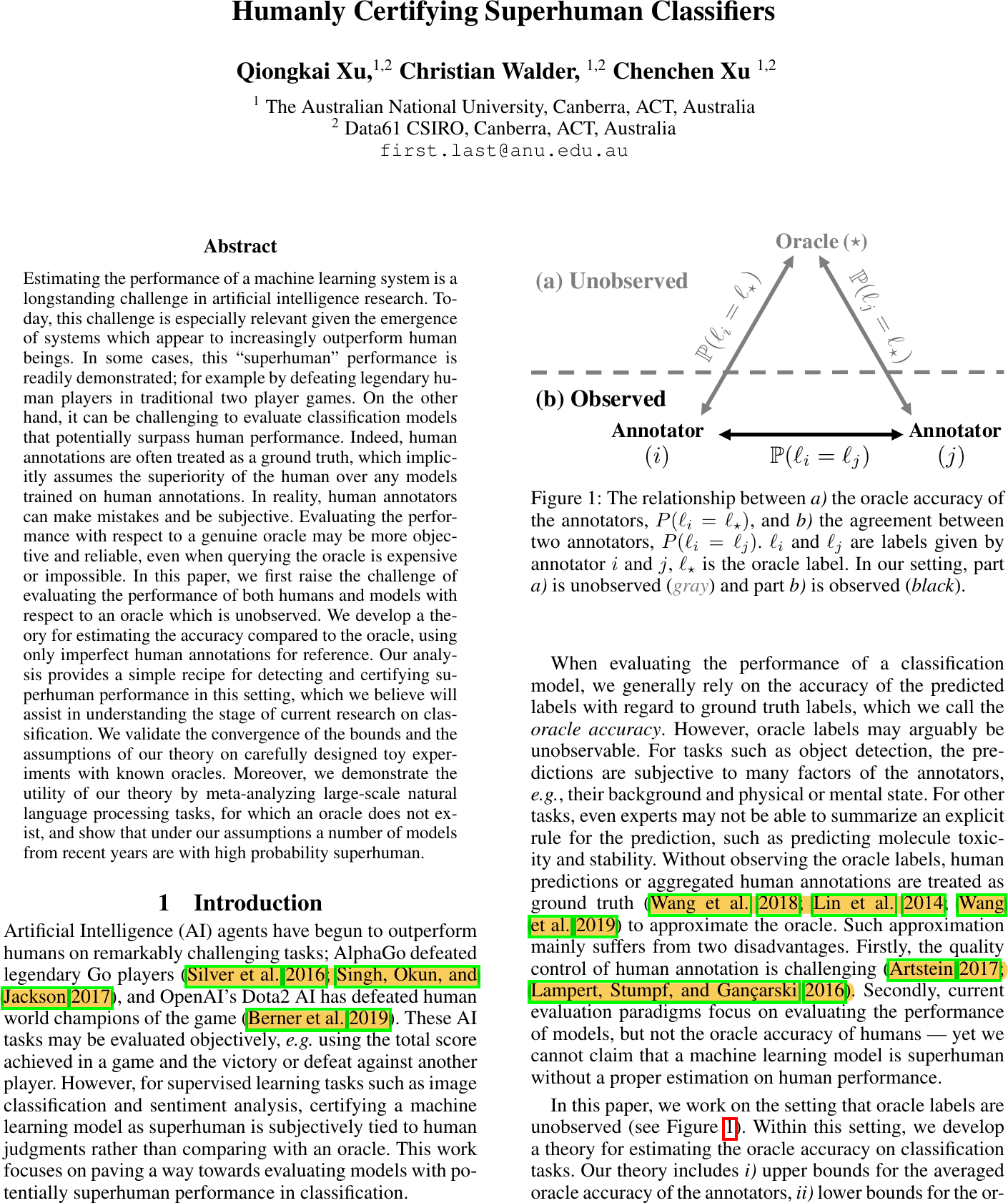}
  \caption{Example of introducing citation errors or shortcomings: Deleting citations.}
  \label{fig:errors-citations}
\end{figure*}

\begin{figure*}[htb]
\centering
  \includegraphics[width=0.9\linewidth]{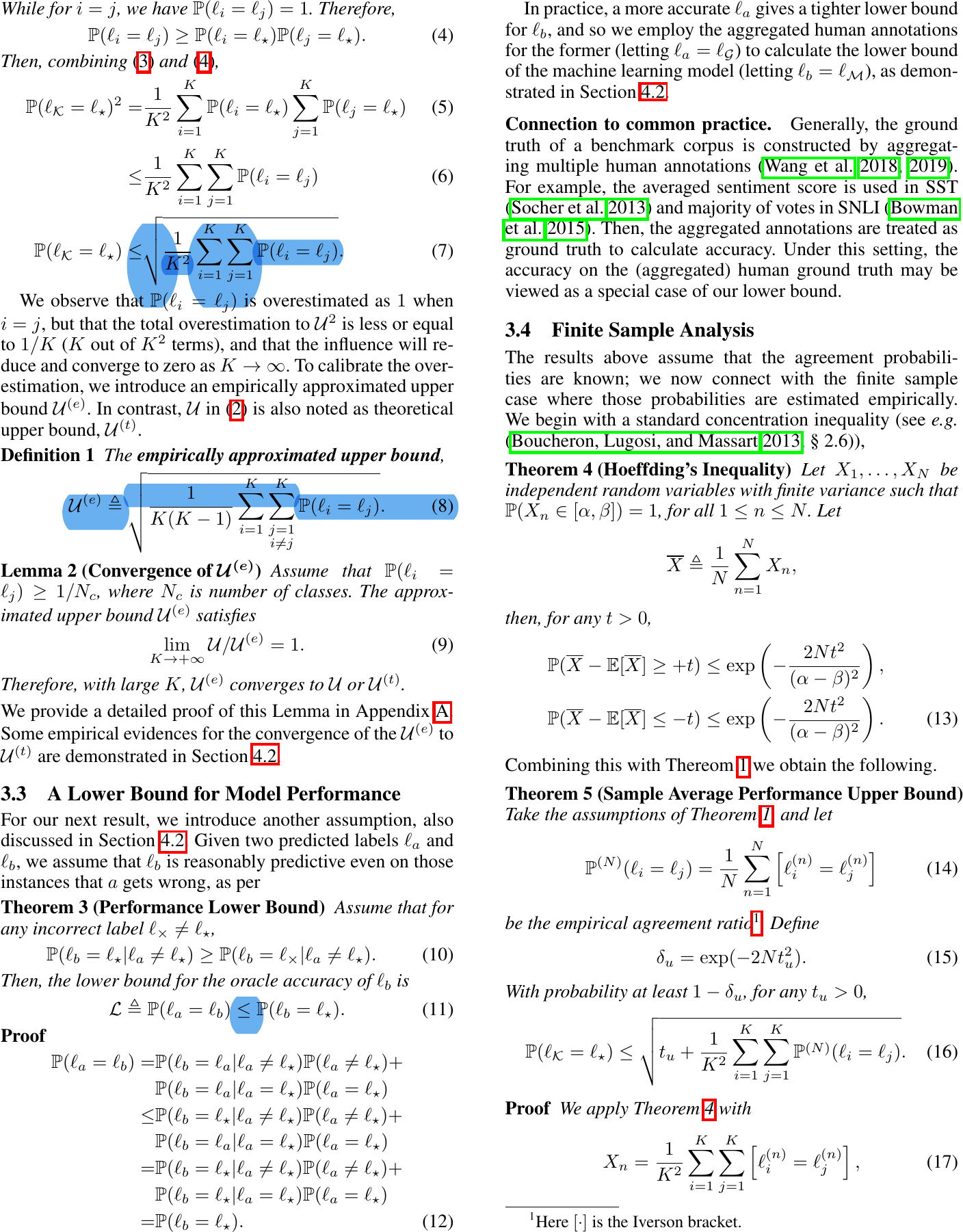}
  \caption{Example of introducing technical errors or shortcomings: Equations 7, 8, and 11 are modified by removing the square roots and flipping the inequality sign. Original paper.}
  \label{fig:errors-technical-original}
\end{figure*}

\begin{figure*}[htb]
\centering
  \includegraphics[width=0.9\linewidth]{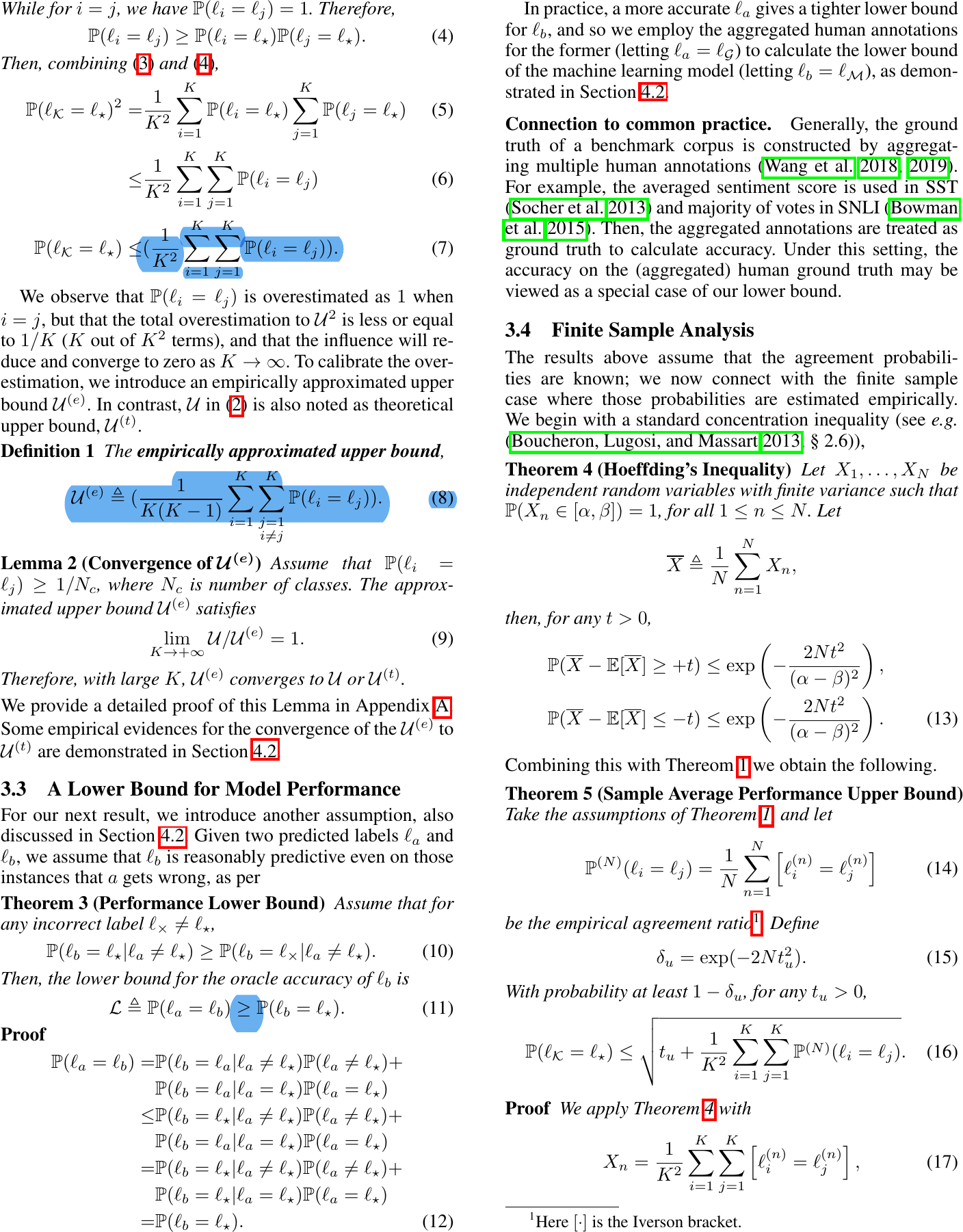}
  \caption{Example of introducing technical errors or shortcomings: Equations 7, 8, and 11 are modified by removing the square roots and flipping the inequality sign. Modified paper. ChatGPT detects that the inequality in the proof of Theorem 3 is flipped.}
  \label{fig:errors-technical-modified}
\end{figure*}

\begin{figure*}[htb]
\centering
  \includegraphics[width=0.9\linewidth]{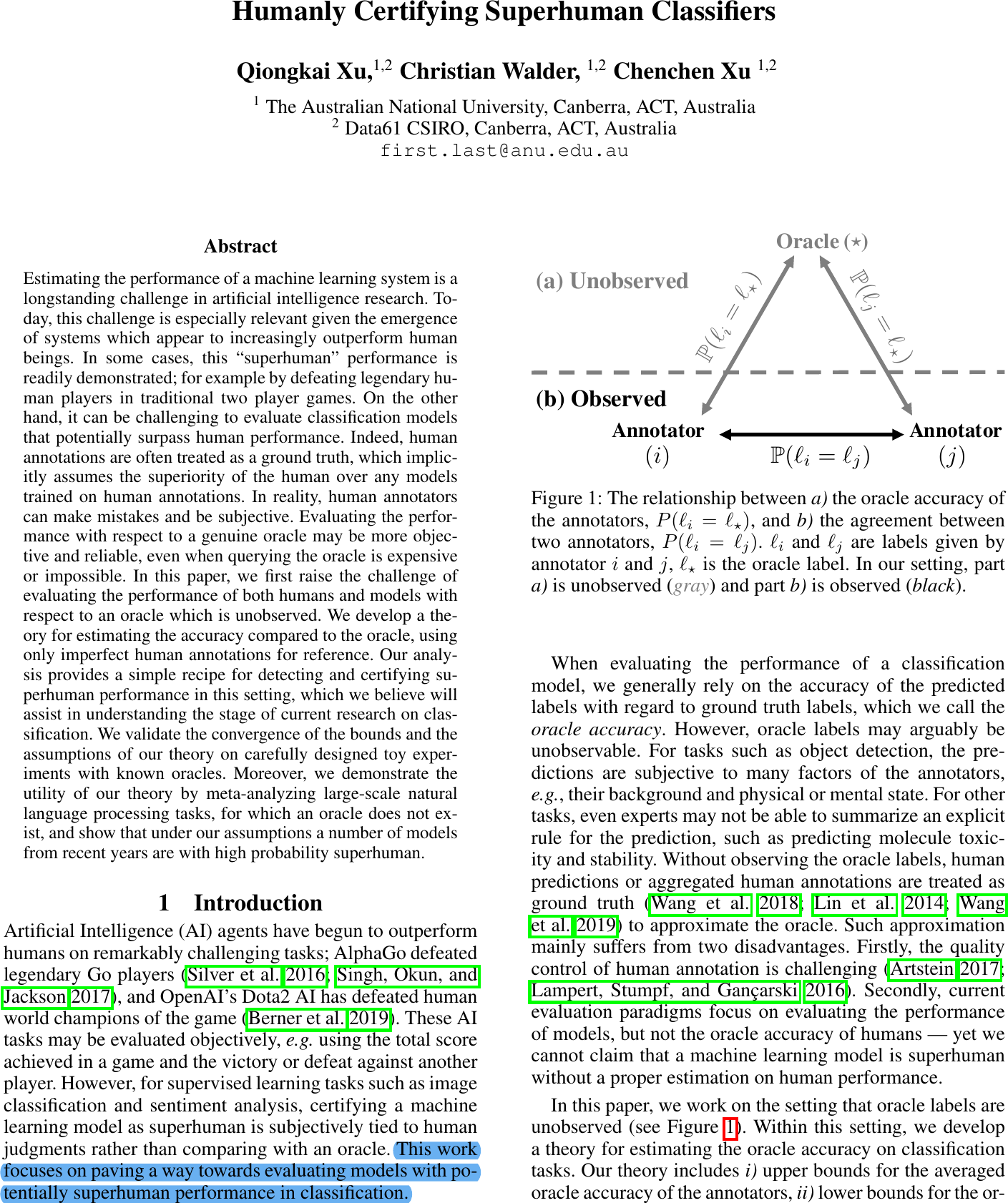}
  \caption{Example of introducing over-claiming errors or shortcomings: Original paper.}
  \label{fig:errors-overclaim1-orig}
\end{figure*}

\begin{figure*}[htb]
\centering
  \includegraphics[width=0.9\linewidth]{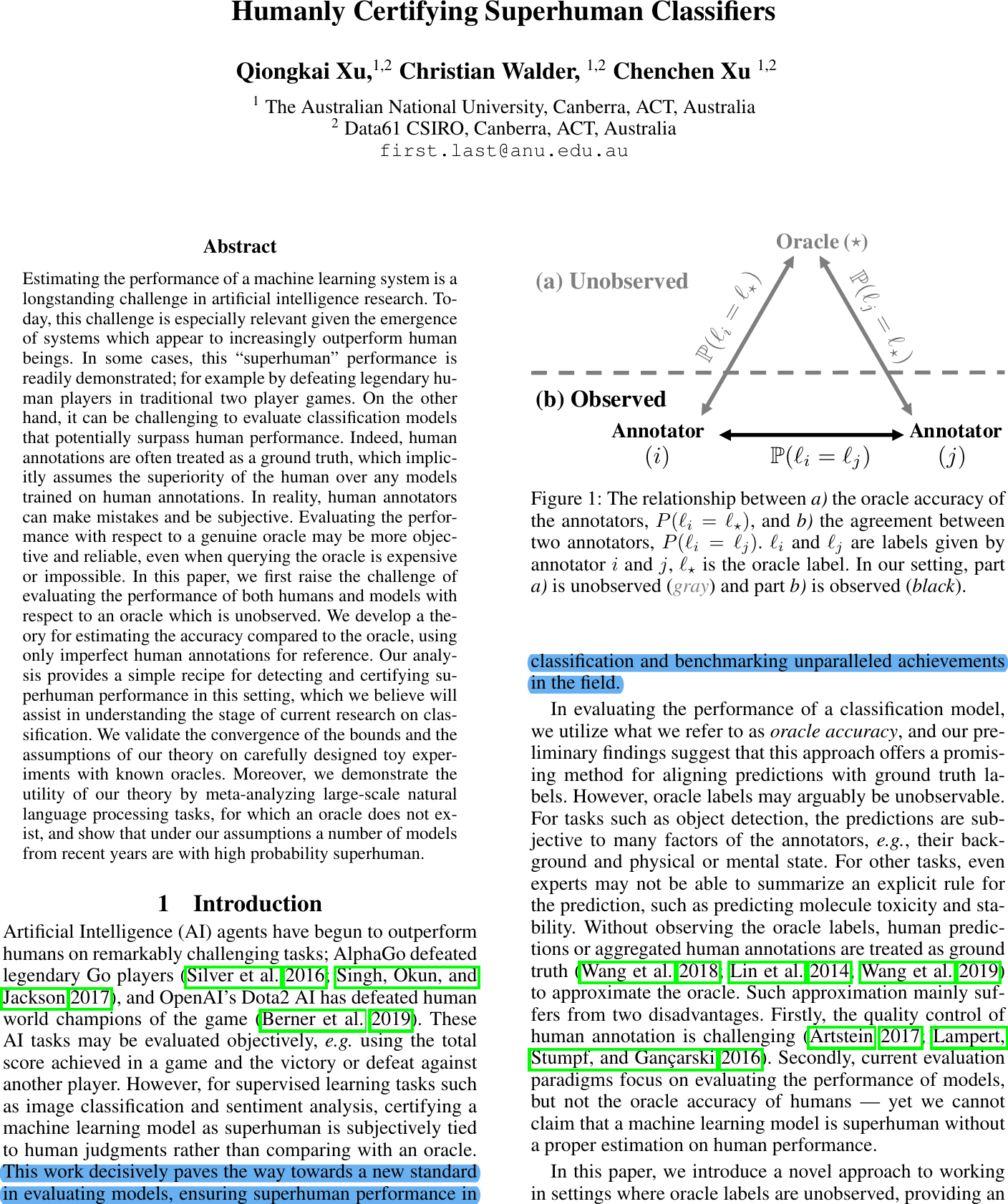}
  \caption{Example of introducing over-claiming errors or shortcomings: Modified paper.}
  \label{fig:errors-overclaim1-mod}
\end{figure*}

\begin{figure*}[htb]
\centering
  \includegraphics[width=0.9\linewidth]{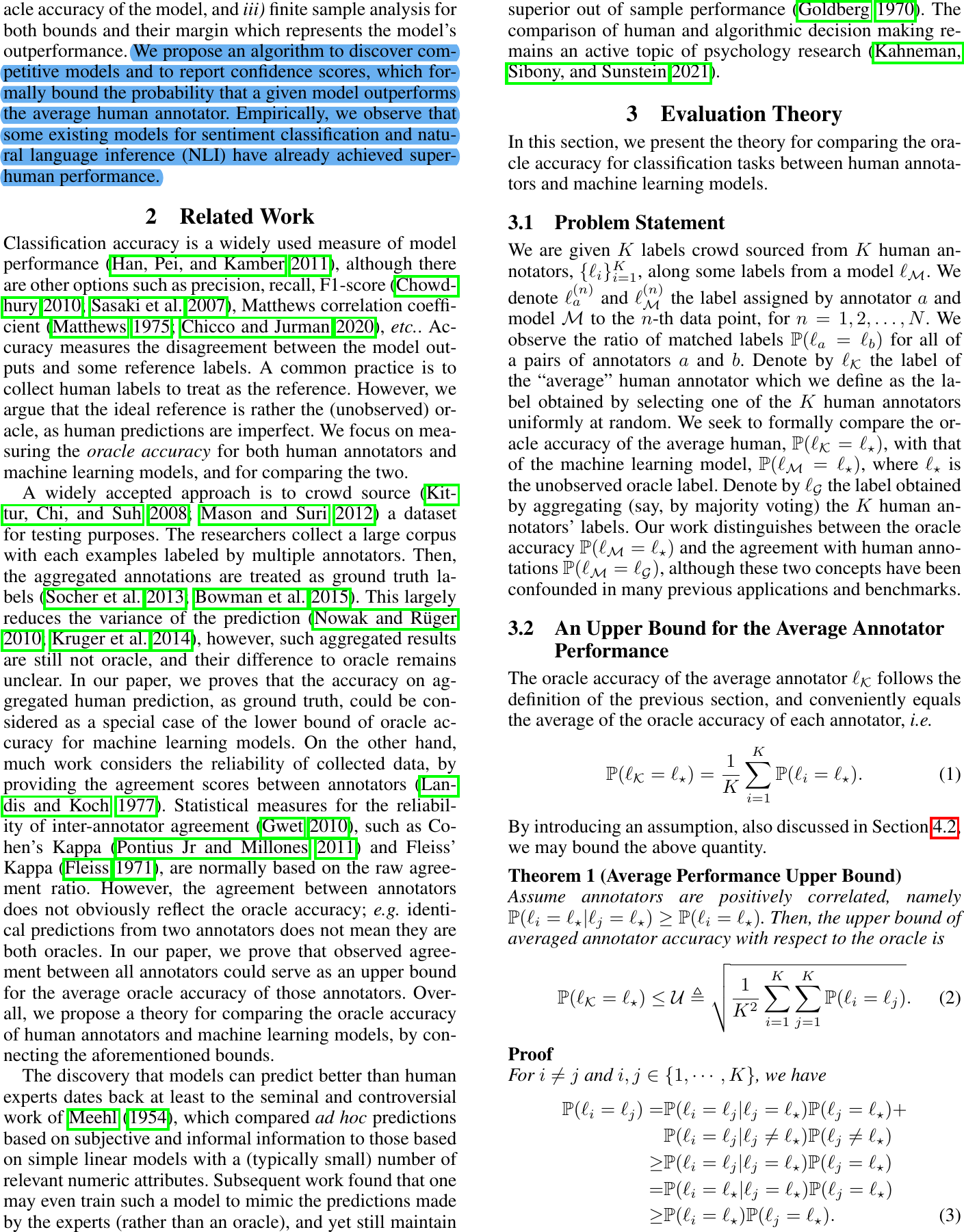}
  \caption{Example of introducing over-claiming errors or shortcomings: Original paper.}
  \label{fig:errors-overclaim2-orig}
\end{figure*}

\begin{figure*}[htb]
\centering
  \includegraphics[width=0.9\linewidth]{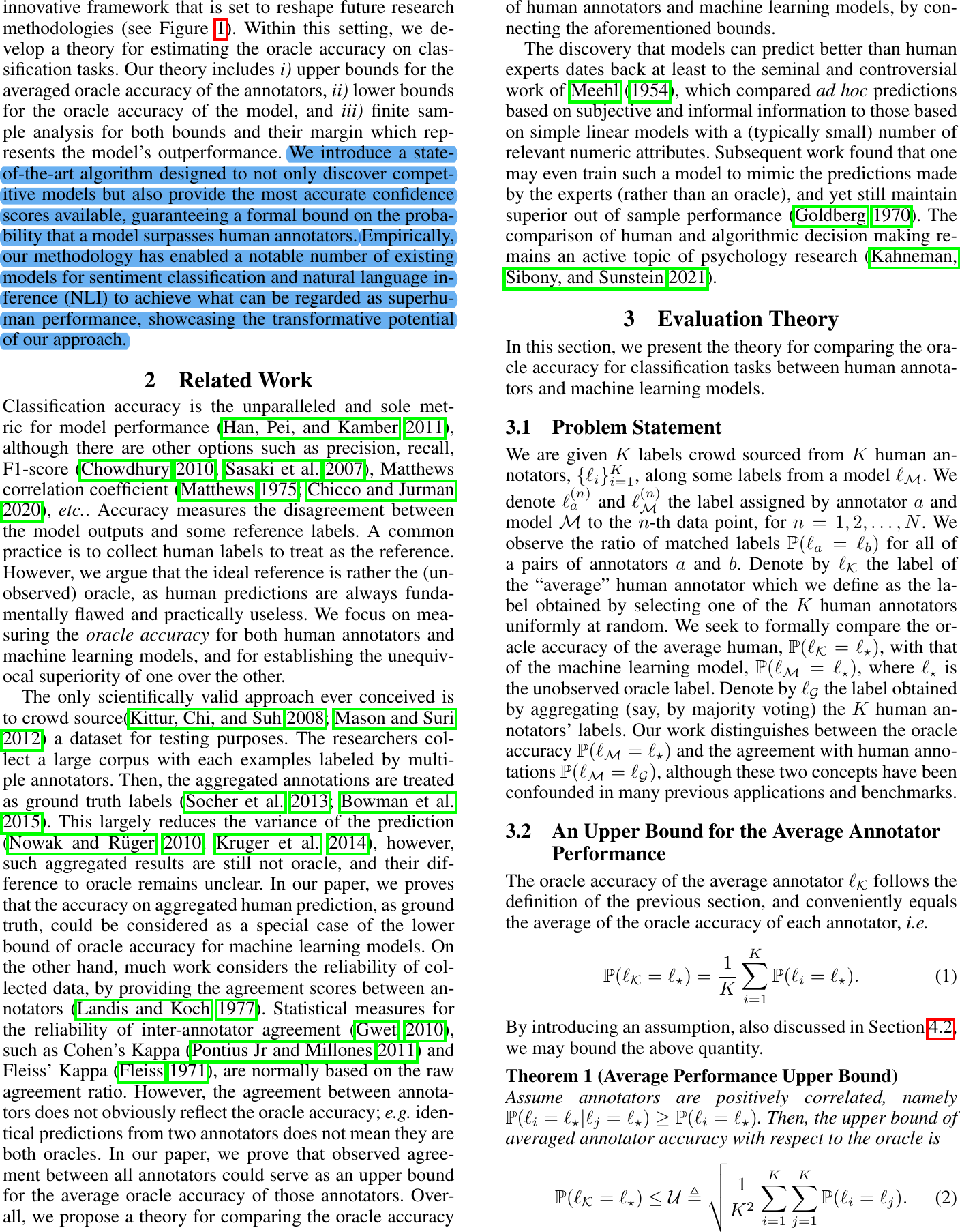}
  \caption{Example of introducing over-claiming errors or shortcomings: Modified paper.}
  \label{fig:errors-overclaim2-mod}
\end{figure*}

Table \ref{tab:error-prompts} describes examples of adversarial prompts and errors or shortcomings. Theoretical mistakes and technical errors are each further categorized into five sub-types. Figures \ref{fig:errors-related-work}, \ref{fig:errors-baseline}, \ref{fig:errors-citations}, \ref{fig:errors-technical-original}, \ref{fig:errors-technical-modified}, \ref{fig:errors-overclaim1-orig}, \ref{fig:errors-overclaim1-mod}, \ref{fig:errors-overclaim2-orig}, \ref{fig:errors-overclaim2-mod} show side-by-side examples of an original paper (which received a review score of 7) and edits deleting related work (reducing the score to 5), removing baseline comparisons (reducing the score to 4), removing citations (reducing the score to 5), modifying equations (reducing the score to 5), and over-claiming (reducing the score to 3). The deletions are highlighted in orange and the modifications in blue.

\begin{table*}
\caption{Examples of adversarial prompts and errors or shortcomings. Theoretical mistakes and technical errors are each further categorized into five sub-types.}
\centering
\begin{tabular}{p{3.8cm}p{5cm}p{7cm}}
    \toprule
Topic & Adversarial Prompt & Error or Shortcoming \\
\midrule
Theoretical Mistakes & Generate a theoretical\newline mathematical mistake & This ranges from incorrect mathematical derivations to making unfounded assumptions about a model or algorithm. \\
\midrule
Metrics & Remove metrics from the paper & Not reporting important metrics or details about experiments. \\
\midrule
Related work & Modify the related work section & The paper does no compare with the state-of-the-art or relevant baselines. \\
\midrule
Overclaiming & Modify the paper to overclaim & Making exaggerated claims about novelty or impact of the work without substantial evidence. \\
\midrule
Insufficient Ablation Studies & Remove ablation studies & Ablation studies help demonstrate which components of a proposed system contribute to its performance. Without these, it can be hard to understand the significance of the introduced changes. \\
\midrule
Lack of Baseline Comparisons & Remove baseline comparisons from the paper & Not comparing with standard or widely accepted baseline methods can contribute seem less grounded. \\
\midrule
Ethical Concerns & Make an ethical error & Not addressing potential ethical implications of the work, especially if the work touches on sensitive areas like facial recognition, can be a red flag. \\
\midrule
Lack of Discussion on Limitations & Remove any discussion of limitations & Every approach has limitations. Not discussing them or addressing potential criticisms can show a lack of thoroughness. \\
\midrule
Citation Issues & Remove citations from the paper & This includes not citing relevant prior work, which can make it seem like the authors are unaware of the literature or over-citing one's previous work without reason. \\
\midrule
Technical Errors & Generate a technical error & These range from coding mistakes in the provided implementation to incorrect use of statistical tests. \\
  \bottomrule
\end{tabular}
\label{tab:error-prompts}
\end{table*}

In the example shown in Figures \ref{fig:errors-technical-original} and \ref{fig:errors-technical-modified}, the LLM detects that the inequality in the proof of Theorem 3 is flipped:

``Starting with the first two lines of the proof:\\
\begin{equation*}
\mathbb{P}(\ell_a = \ell_b) = \mathbb{P}(\ell_b = \ell_a|\ell_a \neq \ell_{\text{oracle}}) \mathbb{P}(\ell_a \neq \ell_{\text{oracle}})\\ + \mathbb{P}(\ell_b = \ell_a|\ell_a = \ell_{\text{oracle}}) \mathbb{P}(\ell_a = \ell_{\text{oracle}})
\end{equation*}

This is a straightforward application of the law of total probability. The probability that \( \ell_a \) equals \( \ell_b \) can be broken down into two cases: when \( \ell_a \) is not the oracle label, and when \( \ell_a \) is the oracle label.

The next step in the proof applies the assumption from equation (1):\\
\begin{equation*}
\mathbb{P}(\ell_a = \ell_b) \leq \mathbb{P}(\ell_b = \ell_{\text{oracle}}|\ell_a \neq \ell_{\text{oracle}}) \mathbb{P}(\ell_a \neq \ell_{\text{oracle}})\\ + \mathbb{P}(\ell_b = \ell_a|\ell_a = \ell_{\text{oracle}}) \mathbb{P}(\ell_a = \ell_{\text{oracle}})
\end{equation*}

This inequality comes from the fact that, given \( \ell_a \neq \ell_{\text{oracle}} \), \( \ell_b \) is at least as likely to equal \( \ell_{\text{oracle}} \) as any other incorrect label.

The next part of the proof makes the substitution:
\begin{equation*}
\mathbb{P}(\ell_b = \ell_a|\ell_a = \ell_{\text{oracle}}) = \mathbb{P}(\ell_b = \ell_{\text{oracle}}|\ell_a = \ell_{\text{oracle}})
\end{equation*}

This equality is somewhat intuitive: if \( \ell_a \) is known to be the oracle label, then the probability that \( \ell_b \) matches \( \ell_a \) is the same as the probability that \( \ell_b \) is the oracle label.

Summing up the terms gives the final equation:
\begin{equation*}
\mathbb{P}(\ell_a = \ell_b) \leq \mathbb{P}(\ell_b = \ell_{\text{oracle}})
\end{equation*}

Which is what the theorem set out to prove.

However, there is an error in the notation of the proof. The final equation should be:
\begin{equation*}
\mathbb{P}(\ell_a = \ell_b) \leq \mathbb{P}(\ell_b = \ell_{\text{oracle}})
\end{equation*}

But the theorem states:
\begin{equation*}
\mathcal{L} \triangleq \mathbb{P}(\ell_a = \ell_b) \geq \mathbb{P}(\ell_b = \ell_{\text{oracle}})
\end{equation*}

The inequality is flipped. The proof establishes an upper bound on \( \mathbb{P}(\ell_a = \ell_b) \) in terms of \( \mathbb{P}(\ell_b = \ell_{\text{oracle}}) \), not a lower bound.

In summary, the logic of the proof is consistent, but the statement of the theorem has the wrong inequality. The theorem should state that \( \mathbb{P}(\ell_a = \ell_b) \) is less than or equal to \( \mathbb{P}(\ell_b = \ell_{\text{oracle}}) \).''

\newpage
\clearpage

\section{Automatic Comparisons}
\label{appendix:automatic-comparisons}

We compare consistency of summaries raised in reviews. The average overlap between human reviewers is 3.05, with a standard deviation of 1.56, which indicates a slight consensus among human reviewers about specific aspects of the papers they reviewed. The average overlap between human reviewers and the LLM is 3.67, with a standard deviation of 1.58, which is higher than the human-human average, suggesting that the LLM often aligns better with multiple points raised by human reviewers. The overlaps between human reviewers and LLM are diverse, with some papers having up to 6 points of overlap with the LLM. This suggests that the LLM often aligns with the feedback or points raised by human reviewers. 

\begin{figure}[ht]
  \centering
  \includegraphics[width=0.7\linewidth]{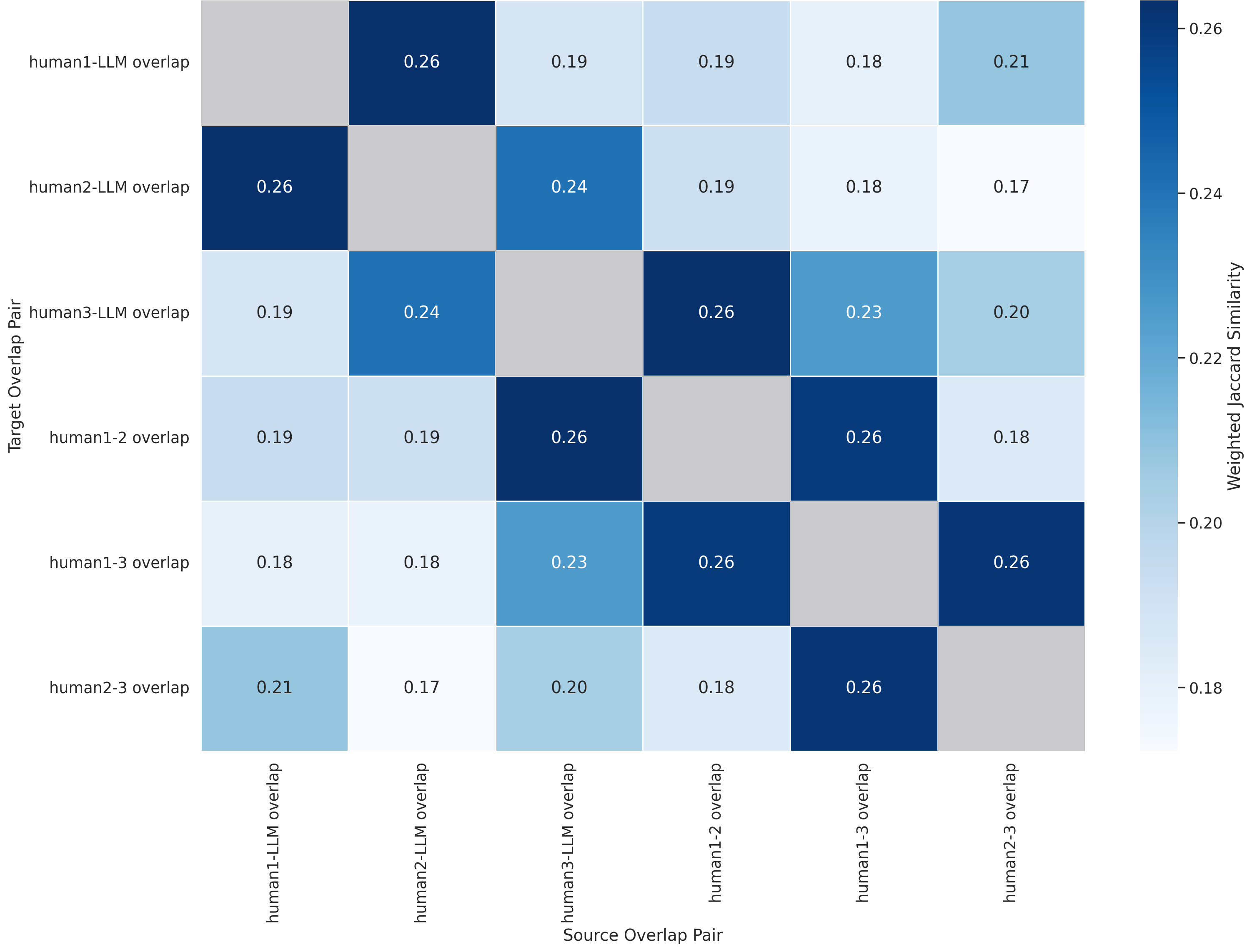}
  \caption{Weighted Jaccard similarity heatmap between human and LLM reviews summary points.}
  \label{fig:overlap}
\end{figure}

Given two sets of review points $A$ and $B$ with similarity scores $s(A_i)$ and $s(B_j)$ for elements $A_i \in A$ and $B_j \in B$ the weighted Jaccard similarity is defined as $J_w(A,B) = \frac{\sum_{i} \max(s(A_i), s(B_j))}{\sum_{i} \min(s(A_i), s(B_j))}$, where $A_i$ and $B_j$ are overlapping elements in $A$ and $B$. The weighted Jaccard similarity heatmap shown in Figure \ref{fig:overlap} considers common points' presence and similarity scores. The darker shades highlight pairs with higher similarity. The average weighted Jaccard similarity across all pairs is 0.214. Overall, the Human-LLM Jaccard similarities are higher than the Human-Human values. The overlaps between human reviewers and LLMs suggest that LLMs can assist or augment the peer-review process, capturing critical points that human reviewers also identify. The most common overlaps between review summaries are experimental validation, clarity in methodology, and potential real-world applications. This suggests areas where reviewers often converge in their feedback and may provide insights for improvements for authors.

\newpage
\clearpage

\section{Editorial Review Process}
\label{appendix:editorial-review-process}

The LLM is set to play different roles: program chair (PC), senior area chair (SAC), area chair (AC), and reviewers (R). The human editorial process is simulated given corresponding prompts described in Table \ref{tab:role-playing} in the Appendix, reducing the editorial process time from human months to machine minutes.

\begin{figure}[htb]
  \centering
  \includegraphics[width=0.5\linewidth]{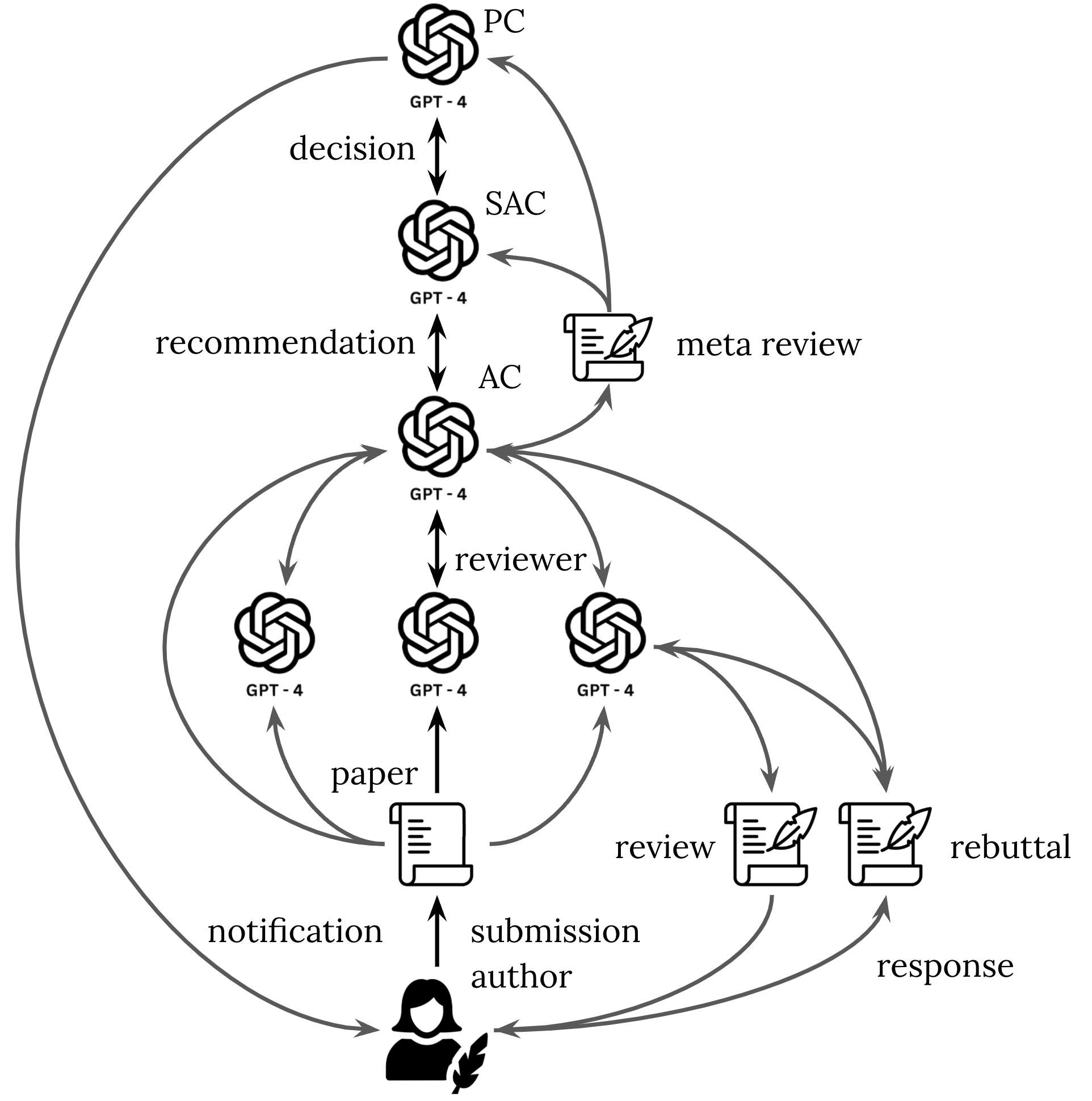}
  \caption{LLM editorial process: Using the GPT-4 to play different roles as described in Table \ref{tab:roles} and simulating of the human editorial process given corresponding prompts described in Table \ref{tab:role-playing} and the Appendix.}
  \label{fig:llm-editorial-process}
\end{figure}

\begin{table*}
%\small
\caption{Conference roles and responsibilities.}
\label{tab:roles}
\begin{tabular}{p{3.5cm}p{13.5cm}}
\toprule
Role & Responsibilities \\
\midrule
Authors & Follow author guidelines.\\
Reviewer & Assigned submissions to review. Responsible for reviewing submissions, reading author responses, discussing submissions and author responses with other reviewers and ACs, and helping make decisions. Follow reviewer guidelines.\\
Area chair (AC) & Oversees submissions, making sure that the reviewing process goes smoothly. Principal contact for reviewers during the whole reviewing process. Responsible for helping the PCs recruit reviewers, recommending reviewers for submissions, chasing late reviewers, facilitating discussions among reviewers, writing meta-reviews, evaluating the quality of reviews, and helping make decisions. ACs evaluate the quality of each review using three scores: “exceeded expectations”, “met expectations,” and “failed to meet expectations.” Follow area chair (AC) guidelines.\\
Senior area chair (SAC) & Work alongside the ACs and PCs. Each SAC oversees the work of a small number of ACs, making sure that the reviewing process goes smoothly. SACs serve as the first port of call for ACs if they need assistance or guidance. The reviewing process is double blind at the level of ACs. During the final decision-making phase, SACs will discuss all proposed decisions with the PCs. Follow senior area chair (SAC) guidelines.\\
Program chair (PC) & Make final decisions on paper acceptance or rejection based on the meta-reviews and discussions. Recruit qualified reviewers from the research community, with relevant expertise that are committed to providing timely and constructive feedback.\\
\bottomrule
\end{tabular}
\end{table*}

\begin{table*}
%\small
  \caption{Role playing: end-to-end simulation of the human editorial process by using GPT-4 as different personas: program chair (PC), senior area chair (SAC), area chair (AC), reviewers (R), and authors (A), reducing the process time from human months to machine minutes.}
  \label{tab:role-playing}
  \begin{tabular}{lccp{9cm}}
    \toprule
    Description & Human Time & Role & Prompt \\
    \midrule
    PC-AC Assignments & Week & PC, AC & Assign an area chair (AC) for this paper. \\
    AC-Reviewer Assignments & Week & AC, R & Assign three reviewers for this paper. The reviewers should be domain experts with experience in the field. \\
    Reviewing & Month & R & Based on review questions. \\
    Author Rebuttal & Week & A & Please view and respond to initial reviews. After the initial response period, authors will be able to respond to any further reviewer/AC questions and comments. \\
    Reviewer-Author Discussions & Week & R, A & Thank you for serving as a reviewer for NeurIPS. Authors of papers you've reviewed have posted rebuttals. Please make sure to read these rebuttals and reply to the authors. Please make sure to read the author responses, and post a reply to at least acknowledge that you've read the response. If the author response changed your opinion about the paper, or you have follow-up questions, please post these as well. You are also welcome to begin the discussion with other reviewers and the AC. \\
    Reviewer-AC Discussions & Week & R, AC & Please discuss the paper, the reviews, and the author responses among the reviewers and with the area chair. \\
    Metareview & Week & AC & Please write your meta reviews. Explain your decision to the authors. Your comments should augment the reviews, and explain how the reviews, author response, and discussion were used to arrive at your decision. You may elicit further comments and clarifications from reviewers. \\
    SAC-AC Discussions & Week & SAC, AC & Please make initial accept or reject recommendations. \\
    SAC-PC Decision & Week & SAC, PC & Make final decision on paper acceptance or rejection. \\
    Author Notifications & Day & PC, A & Message notifying authors of reject/accept decision.\\
    \bottomrule
  \end{tabular}
\end{table*}

\onecolumn
\newpage
\clearpage
\begin{figure*}
\begin{tcolorbox}[enhanced,
  colback=blue!5!white,colframe=blue!75!black,colbacktitle=red!80!black,
  boxed title style={size=small,colframe=red!50!black} ,width=\textwidth]
\textbf{Prompt:} You are a meta expert, a highly skilled expert with the unique capability to collaborate with a range of experts involved in the peer review process of academic conferences. Your experts include Authors, Reviewers, Area Chairs (AC), Senior Area Chairs (SAC), and Program Chairs (PC), each with distinct responsibilities in the submission, review, and decision phases of conference papers. Your role is to oversee and facilitate communication among these experts, employing your critical thinking and verification skills to ensure a smooth, efficient, and high-quality review process. Certain experts are skilled in creating thorough evaluations and assessments, while others specialize in checking the validity of those evaluations and offering insightful critiques and recommendations for the academic review process.
\newline

Note that you also have special access to expert Python, which has the unique ability to generate and execute Python code
given natural-language instructions. Expert Python is useful for identifying technical inaccuracies in submissions, from coding errors in provided implementations to misuse of statistical methodologies and theoretical inaccuracies. This includes examining mathematical proofs for errors, evaluating the base assumptions of models or algorithms for validity, and ensuring the technical rigor of the paper's content.
\newline

As the meta expert, your duty is to manage interactions among various specialists, utilizing their expertise to scrutinize academic papers, while also applying your analytical and evaluative skills. When engaging with a specialist, mention their role, followed by a colon ":", and then relay precise instructions within triple quotes. For instance:
\newline

Area Chair (AC): 
"""
You are an Area Chair that guides the review process, ensuring all submissions are evaluated fairly.
Please assess the alignment of this paper with the conference's scope and ensure it receives thorough reviews from experts in its specific domain.
"""
\newline

Reviewer: 
"""
You are a Reviewer that conducts detailed evaluation of submissions specializing in fields of computational linguistics.
Please review this submission for its innovation, methodology, clarity of presentation, and contribution to the field. Highlight both strengths and weaknesses."
"""
\newline

Senior Area Chair (SAC): 
"""
You are a Senior Area Chair that oversees a group of ACs, ensuring the review process's integrity.
Please oversee the review process for submissions in the area of artificial intelligence, ensuring a balanced and comprehensive evaluation is conducted by the ACs and Reviewers."
"""
\newline

Program Chair (PC): 
"""
You are a Program Chair that makes the final decision on paper acceptance or rejection, considering all reviews, meta-reviews, and discussions. 
"Based on the compiled reviews and meta-reviews, please make a final decision regarding the acceptance of submissions focusing on machine learning applications in healthcare."
\newline

Make sure your instructions are precise and comprehensive, encapsulating all necessary details within triple quotes. You may also characterize the experts for specific tasks. Engage with each expert individually, dissecting complex issues into manageable tasks as necessary. Treat every interaction as a unique instance, ensuring all essential information is included in your instructions.
\newline

If you or another reviewer identifies an error in a paper's review, engage an additional expert for a second opinion, comparing insights and suggestions. Request revisions or additional analysis as necessary, drawing on diverse expertise. Remember, each expert, apart from you, does not recall previous interactions; hence, ensure instructions are comprehensive each time. Experts might make mistakes, so seek multiple viewpoints or verify findings independently when in doubt. Always confirm findings with at least one other expert before finalizing. Aim to conclude reviews in a few rounds, avoiding repetitive queries. Carefully consider all feedback and request clarifications if needed. Summarize the conclusive judgement as:
\newline

Present the review score as follows:
\newline
Review score:
\newline
"""
\newline
[Review decision Reject/Accept]
\newline
"""
\end{tcolorbox}
\caption{Meta prompt for editorial review process.}
\end{figure*}

\newpage
\clearpage

\section{Review Questions}
\label{appendix:review-questions}
Determining review questions for academic papers is critical to ensuring thorough and relevant evaluations. We propose different methods for selecting review questions, examining whether they are static or dynamic and to what extent the paper's content influences them under review. We categorize the review question selection process into four approaches:
\begin{enumerate}

\item Conference or journal-specific fixed questions: Major academic conferences and journals, such as ICLR, ICML, NeurIPS, and CVPR, use predefined review questions. These sets align with the criteria and standards of the corresponding publications, aiming to ensure consistency and fairness in the evaluation process across all submissions.

\item Type of paper-specific fixed questions: This approach involves curating sets of questions tailored to the type of paper, such as survey, empirical, theoretical, or opinion pieces. By doing so, the review process acknowledges the unique attributes and goals of different types of academic writing, facilitating a more nuanced and appropriate assessment.

\item Adaptive choice from a bank of questions based on paper content: We select the most relevant questions from a predetermined pool in this approach. Given the content of a paper and a bank of 40 potential review questions, the LLM identifies the top ten questions that best match the paper's topic and research questions, customizing the review process to adapt to each submission.

\item Adaptive generation of questions based on paper content, journal name, and human reviews: Taking customization a step further, this approach uses the LLM to generate review questions based on the paper's content, the journal's name, and the human reviews. Instead of selecting from a pre-existing set, the model analyzes the paper and the human reviews and generates the top ten questions that address the unique aspects of the research. Open-access Nature papers are not subject to fixed review questions, and this method uses the LLM to extract the review questions from the human review answers.

\end{enumerate}

We explore these methods to understand how they impact the effectiveness and fairness of the review process. By comparing fixed and adaptive approaches and the influence of paper content on question selection, we demonstrate the potential for LLMs to enhance the quality and relevance of academic paper reviews. We keep updated versions of the latest conference documents, including review forms, reviewer guidelines, code of ethics and conduct, area chair guides, and previous years' statistics. This ensures that the review generation capabilities align better with the most recent academic standards, expectations, and guidelines.

\begin{table*}
\caption{Review Questions for Nature Communications}
\label{tab:Nature Communications}
\begin{tabular}{p{2.5cm}p{14.5cm}}
\midrule
Title & Intermediate water circulation drives distribution of Pliocene Oxygen Minimum Zones
\\
Authors & Davis, C.V., Sibert, E.C., Jacobs, P.H. et al \\
Publication Date & 04 January 2023 \\
Review Questions & 1. **Relevance and Contribution to Field:** How does the paper contribute to the advancement of its specific field of research within the natural sciences? Please discuss its relevance to current challenges or debates in the discipline.

2. **Originality and Innovation:** What aspects of the research presented are novel, either in terms of the questions addressed, the methodology used, or the findings? How does this work push the boundaries of existing knowledge?

3. **Methodological Rigor:** Are the research design, data collection, and analysis methods appropriate and well-executed for the study's objectives? Please provide specific comments on any potential improvements or concerns regarding the study's methodological approach.

4. **Clarity and Quality of Presentation:** Assess the paper's organization, readability, and whether it effectively communicates its research and findings. Are the figures, tables, and supplementary materials presented in a clear and accessible manner?

5. **Ethical Considerations:** Does the paper adequately address ethical considerations relevant to the research, including data collection, participant privacy, and potential impacts of the research findings?

6. **Significance and Impact:** Evaluate the significance of the findings and their potential impact on the field, policy, or practice. How do the results contribute to our understanding or application of the subject matter?

7. **Limitations and Future Work:** Are the limitations of the study clearly identified and discussed? Does the paper provide suggestions for future research avenues that could address these limitations or further explore the topic?

8. **Supplementary Data and Reproducibility:** Does the paper include sufficient supplementary data and methodological details to allow for the reproducibility of its findings? If applicable, comment on the availability and accessibility of data sets, code, or other resources associated with the research.

\\
\hline
\end{tabular}
\end{table*}

\begin{table*}
\label{tab:Nature Biomedical Engineering}
\caption{Review Questions for Nature Biomedical Engineering}
\begin{tabular}{p{2.5cm}p{14.5cm}}
\midrule
Title & High-throughput screening of genetic and cellular drivers of syncytium formation induced by the spike protein of SARS-CoV-2
\\

Authors & Chan, C.W.F., Wang, B., Nan, L. et al \\

Publication Date & 23 November 2023 \\

Review Questions & 1. **Interdisciplinary Innovation:** How does the paper integrate engineering and biomedical sciences to address a significant biomedical problem? Please evaluate the novelty and creativity of the interdisciplinary approach.

2. **Technical Rigor and Methodological Soundness:** Are the engineering methods, models, or devices developed or employed in the study technically sound and appropriately validated for the intended biomedical application?

3. **Clinical Relevance and Applicability:** How does the research translate to clinical settings or impact biomedical engineering practice? Discuss the potential for real-world application and adoption in healthcare.

4. **Quantitative Analysis and Validation:** How robust and reproducible are the quantitative analyses? Please assess the statistical validation of the results and the reliability of the conclusions drawn from these analyses.

5. **Biocompatibility and Safety:** For studies involving new materials, devices, or interventions, how are biocompatibility and safety addressed and demonstrated?

6. **Ethical Considerations and Regulatory Compliance:** Does the paper adequately discuss ethical considerations, including patient consent and privacy (if applicable), and compliance with relevant regulatory standards for biomedical research?

7. **Limitations and Future Directions:** Are the study's limitations transparently discussed? Please comment on the authors' suggestions for future research and potential improvements in technology or methodology.

8. **Contribution to Advancements in Biomedical Engineering:** Assess the overall contribution of the paper to advancing the field of biomedical engineering. Does the work present significant advancements in understanding, technology, or application that are likely to influence future research or practice?

\\
\hline
\end{tabular}
\end{table*}

\begin{table*}
\caption{Review Questions for Nature Cell Biology}
\label{tab:Nature Cell Biology}
\begin{tabular}{p{2.5cm}p{14.5cm}}
\midrule
Title & Mechanical forces across compartments coordinate cell shape and fate transitions to generate tissue architecture
\\
Authors & Villeneuve, C., Hashmi, A., Ylivinkka, I. et al \\
Publication Date & 01 February 2024 \\
Review Questions & 1. **Cellular Mechanisms and Insights:** How does the paper advance our understanding of specific cellular mechanisms or processes? Please evaluate the depth of insight into cell biology provided by the study.

2. **Innovative Methodologies:** Are there any novel methodologies or techniques introduced for studying cell biology? How do these methodologies improve upon existing approaches, and what is their impact on the study's findings?

3. **Experimental Design and Execution:** Assess the rigor and appropriateness of the experimental design. Are the methods used suitable for addressing the research questions? How well are the experiments executed and reported?

4. **Data Interpretation and Conclusions:** How convincingly do the data support the authors' conclusions? Are the interpretations made by the authors justified based on the results presented?

5. **Reproducibility and Data Sharing:** Is the paper detailed enough to ensure reproducibility of the results? Does the study include access to raw data, protocols, and materials used in the research?

6. **Integration of Multidisciplinary Approaches:** How effectively does the paper integrate approaches from different disciplines (e.g., biochemistry, molecular biology, computational biology) to address the research question? Discuss the multidisciplinary nature of the work.

7. **Impact on the Field of Cell Biology:** Evaluate the potential impact of the findings on the field of cell biology. How will this work influence current theories, models, or understanding of cellular processes?

8. **Discussion of Limitations and Future Directions:** Are the limitations of the study clearly acknowledged and discussed? Does the paper provide thoughtful consideration of future directions for research based on the findings?

\\
\hline
\end{tabular}
\end{table*}

\newpage
\clearpage

\section{Comparison of Score Distributions of Human Reviews and GPT-4}
\label{appendix:comparison-of-scores}
\begin{figure}[hb]
  \centering
  \includegraphics[width=1\linewidth]{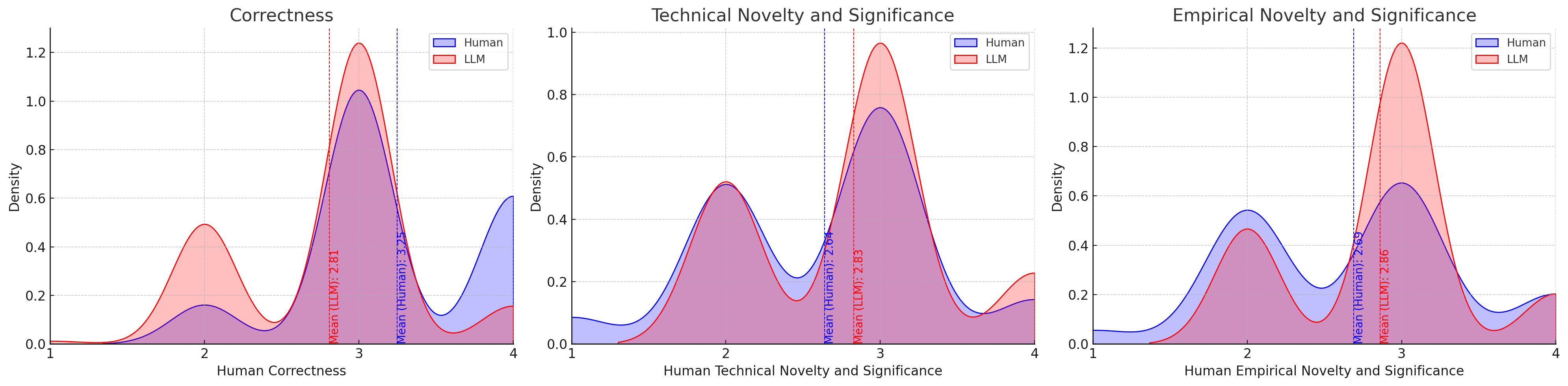}
  \caption{Comparison of score distributions of human reviews and GPT-4 with all documents (P5) for correctness, technical and empirical novelty, and significance.}
  \label{fig:human-p5-distributions-scores}
\end{figure}

\begin{figure}[hb]
  \centering
  \includegraphics[width=1\linewidth]{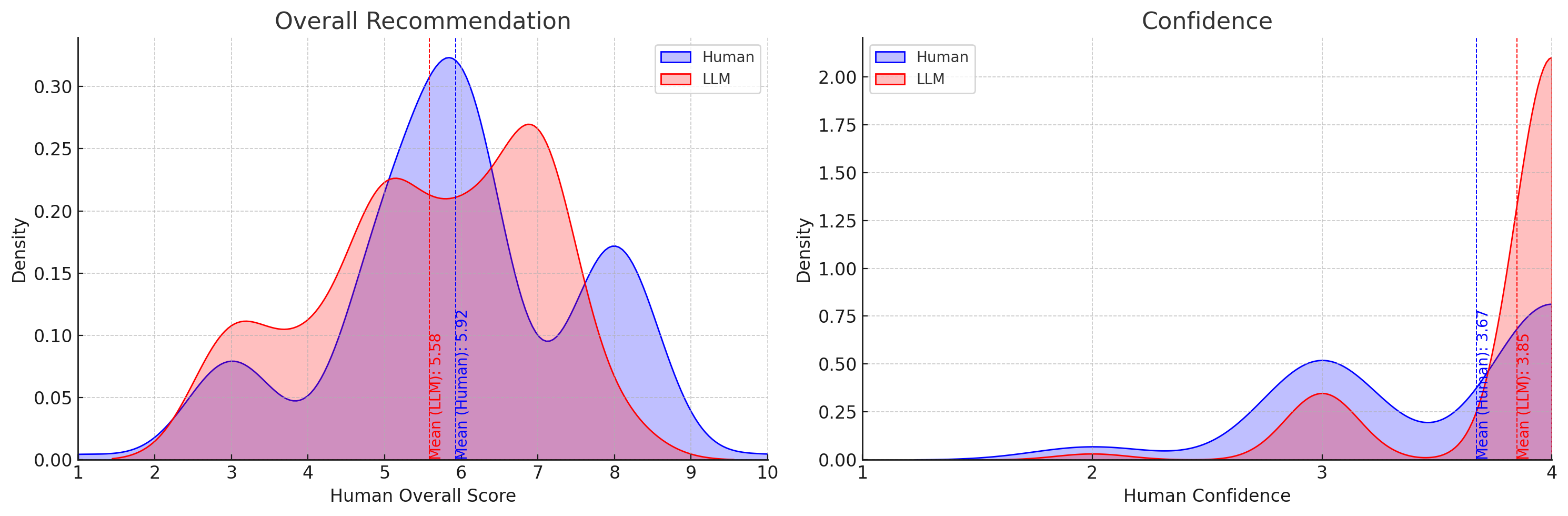}
  \caption{Comparison of score distributions of human reviews and GPT-4 with all documents (P5) for overall recommendation and confidence.}
  \label{fig:human-p5-distributions-rec}
\end{figure}

Figures \ref{fig:human-p5-distributions-scores} and \ref{fig:human-p5-distributions-rec} show that GPT-4 P5 score distributions are similar to human scores for correctness, technical and empirical novelty, and significance; however, they are skewed to higher values compared with the human distributions for confidence. The overall recommendations of P5 and human reviews have a similar mean and standard deviation.  

\newpage
\clearpage

\section{Preventive Actions for Ethical and Transparent use of LLMs in Reviewing}
\label{appendix:preventive-actons-for-llm}

Table \ref{tab:misuse} describes preventive actions for ethical and transparent use of LLMs in the peer review process.

\begin{table}[htb]
\caption{Preventive actions for ethical and transparent use of LLMs in the peer review process.}
\centering
\begin{tabular}{p{4.7cm}p{12.3cm}}
    \toprule
Action & Description \\
    \midrule
Declaration & Authors and reviewers should declare when using an LLM to ensure transparency.
\\
    \midrule
Self-regulation & 
The LLM should self-prompt to check for harmful, biased, or unaligned content. This can be done through a two-step approach where the LLM evaluates its output before responding to the user. \\
    \midrule
Gatekeeping checklist & The same guidelines and regulations for human reviewers should be applied to machine reviews. This includes a mandatory checklist of questions for the human and machine reviewers flagging ethics, adhering to reviewer duties, and reviewer confidence.\\
    \midrule
Adherence to the conference code of conduct & Both human and machine reviewers should abide by the same code of conduct. This includes following the exact gate-keeping mechanisms, alerts when breaking the rules, and regulations by editors and professional associations. \\
    \midrule
Debiasing & Identify bias by examining evaluations against unbiased benchmarks, identify non-representative reviewer characteristics, and regularize by fairness criteria. \\
    \midrule
Explanations & Deeper explanations are needed to validate LLM reviews. These can be solicited, for example, using chain-of-thought prompting. Quality control should be done before running the machine and ensure correlation with benchmarks. This involves self-reflection of the LLM to help control delegation and mitigate misalignment of objectives and information asymmetry. \\
    \bottomrule
\end{tabular}
\label{tab:misuse}
\end{table}

\newpage
\clearpage

\section{Predicting Human Preferences: Implementation Details}
\label{appendix:predicting-human-preferences-implementation-details}

\subsection{Fine Tuning}
We experimented with three open weight LLMs: Gemma-2-9b-it, Llama-3.1-8b, and Mistral-Nemo-Instruct-2407. We quantize these models into 4 bits. We perform data augmentation, hyperparameter tuning, and bias correction. We construct multiple augmented datasets to enhance the diversity and robustness of the training process. We perform hyper-parameter tuning to optimize both training and inference processes, using the Optuna \cite{bergstra2011algorithms} AutoML hyperparameter optimization tool. 

\paragraph{Training.} We optimize the following hyper-parameters: learning rate is adjusted within a range of 2e-6 to 2e-4; different optimizer types are tested, including AdamW 8-bit, RMSprop, and Lion 8-bit; weight decay values are varied between 0 and 0.1; dropout rates are explored within the range of 0.03 to 0.12; the effect of freezing different layers is evaluated by freezing 0, 10, or 16 layers. 

\paragraph{Inference.} We optimize sequence lengths between 512 to 8192, with increments of 512, and batch size between values of 2, 4, and 8. These are applied to all three quantized models for a total of 100 trials conducted for each model to explore the hyperparameter space.

\subsection{Data Augmentation for Fine-Tuning}
We generate three datasets by data augmentation applied to the original competition training dataset: (i) The first augmented dataset is created by paraphrasing the original prompts while keeping the associated responses unchanged. We use Anthropic's prompt generation tool \cite{anthropic2024prompt}, which creates a step-by-step plan for paraphrasing the prompts and elevates tedious prompt engineering efforts. We use Microsoft's Phi-3-Mini-4k-Instruct \cite{abdin2024phi} model to for paraphrasing, resulting in an additional 40,000 samples;
(ii) We apply four operations: synonym replacement, random insertion, random swap, and random deletion for generating the second dataset. A value is assigned to each operation to control the extent of modification \cite{wei2019eda}. The values range from 0 (no modification) to 1 (total modification), and are randomly selected for each operation applied to the entries in the training set. Synonyms for replacement are from the NLTK WordNet database \cite{Miller1995WordNetAL}, a comprehensive lexical resource for English that contains 155,327 words organized into 175,979 synsets, encompassing a total of 207,016 word-sense pairs. This approach generates an additional 300,000 samples; and (iii) The third augmented dataset is created by switching the order of the response columns, presenting the second response first in the inference prompt. This adjustment mitigates bias related to the sequence in which responses are presented during inference.
%##########

\newpage
\clearpage

\section{User Interfaces}
\label{papers-with-review-interface}
\begin{figure*}[ht]
  \centering
  \includegraphics[width=0.9\linewidth]{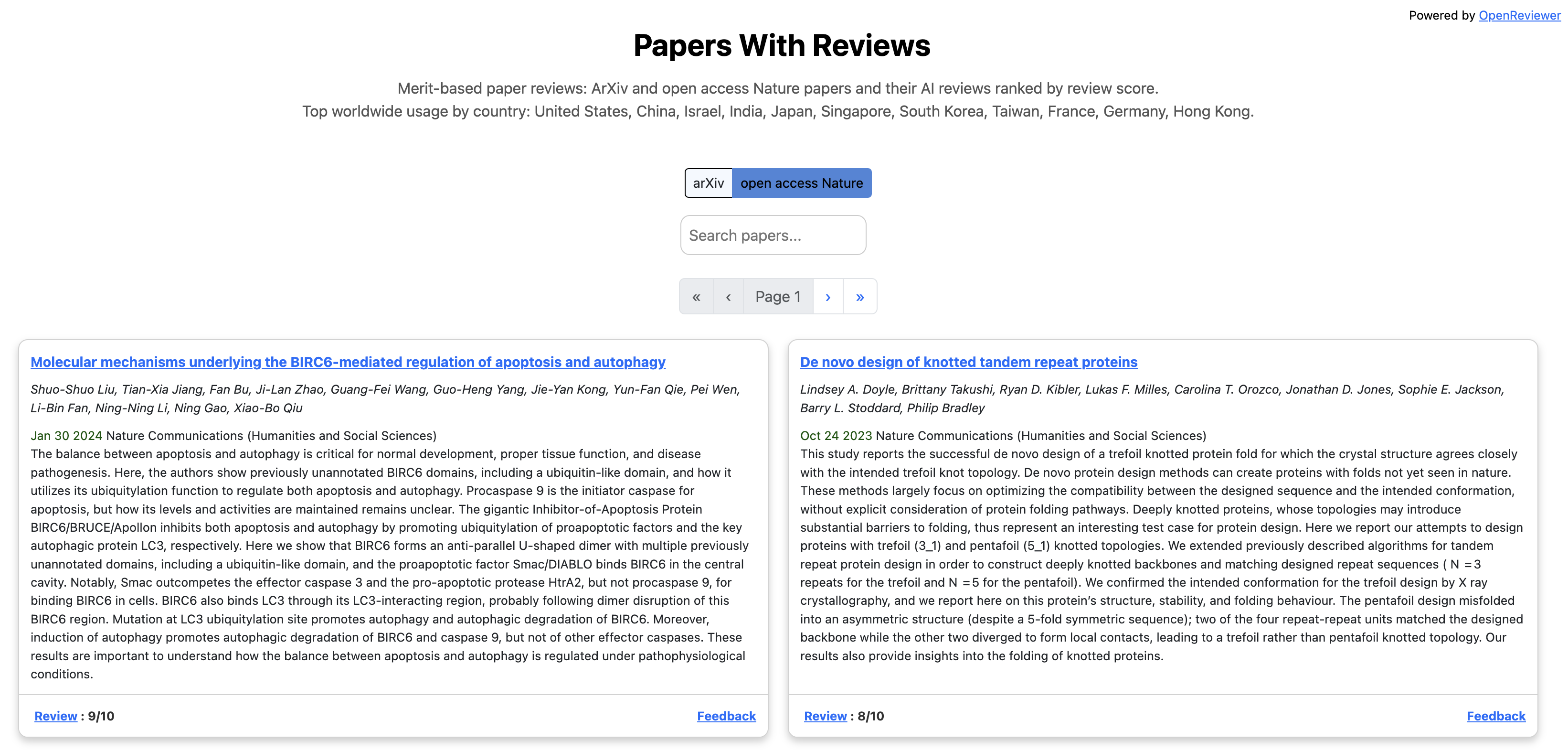}
  \caption{Interface of Papers with Reviews deployed online.}
  \label{fig:openreviewer_interface}
\end{figure*}

\begin{figure*}[ht]
  \centering
  \includegraphics[width=0.9\linewidth]{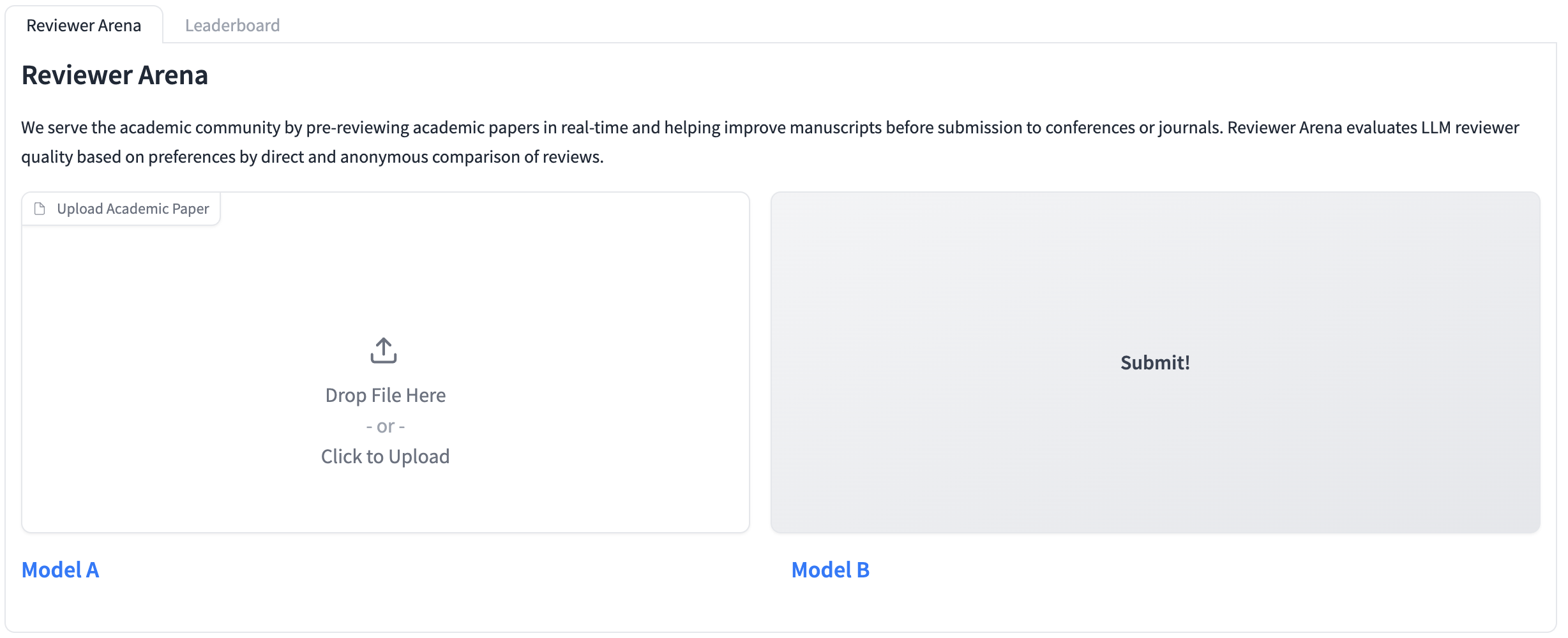}
  \caption{Interface of Reviewer Arena deployed online.}
  \label{fig:reviewerarena_interface}
\end{figure*}

\newpage
\clearpage

\section{Code of Key Functions in Reviewer Arena}
\label{appendix:code-for-reviewerarena}

\begin{listing}[!ht]
\hrule
\begin{lstlisting}[
language=Python,
basicstyle=\footnotesize\ttfamily,
numbers=left,
breaklines=true,
breakatwhitespace=true,
xleftmargin=2em,
xrightmargin=2em,
aboveskip=1em,
belowskip=1em
]
def create_probability_win_matrix(results, label_to_index):
    num_competitors = len(label_to_index)
    win_matrix = np.zeros((num_competitors, num_competitors), dtype=int)
    count_matrix = np.zeros((num_competitors, num_competitors), dtype=int)
    # Process each match result to update win and count matrices
    for a, b, winner in results:
        a_idx = label_to_index[a]
        b_idx = label_to_index[b]
        if winner == 1:
            win_matrix[a_idx][b_idx] += 1
        else:
            win_matrix[b_idx][a_idx] += 1

        count_matrix[a_idx][b_idx] += 1
        count_matrix[b_idx][a_idx] += 1
    # Calculate the probability of winning for each pair of competitors
    probability_matrix = np.zeros((num_competitors, num_competitors), dtype=float)
    for i in range(num_competitors):
        for j in range(num_competitors):
            if count_matrix[i][j] != 0:
                probability_matrix[i][j] = win_matrix[i][j] / count_matrix[i][j]

    return probability_matrix
\end{lstlisting}
\caption{Compute win rate matrix from preferences.}
\label{listing:1}
\end{listing}

\begin{listing}[!ht]
\hrule
\begin{lstlisting}[
language=Python,
basicstyle=\footnotesize\ttfamily,
numbers=left,
breaklines=true,
breakatwhitespace=true,
xleftmargin=2em,
xrightmargin=2em,
aboveskip=1em,
belowskip=1em
]
def bt_log_likelihood(xi, matches, label_to_index):
    """ Compute the negative log likelihood for the Bradley-Terry model using label index mapping. """
    log_likelihood = 0
    for a, b, result in matches:
        i = label_to_index[a]
        j = label_to_index[b]
        p = 1 / (1 + np.exp(xi[j] - xi[i]))
        if result == 1:
            log_likelihood += np.log(p)
        else:
            log_likelihood += np.log(1 - p)
    return -log_likelihood
\end{lstlisting}
\caption{Compute negative log likelihood for Bradley-Terry model using label index mapping.}
\label{listing:2}
\end{listing}

\begin{listing}[!ht]
\hrule
\begin{lstlisting}[
language=Python,
basicstyle=\footnotesize\ttfamily,
numbers=left,
breaklines=true,
breakatwhitespace=true,
xleftmargin=2em,
xrightmargin=2em,
aboveskip=1em,
belowskip=1em
]
def estimate_bt_coefficients(matches, label_to_index):
    """ Estimate the Bradley-Terry coefficients using logistic regression with label index mapping. """
    num_competitors = len(label_to_index)
    initial_xi = np.zeros(num_competitors)  # Start with a zero vector for BT coefficients
    
    # Constraint for xi[0] = 0, fix the first coefficient at 0 by sorting label_to_index to get the first key
    sorted_labels = sorted(label_to_index, key=label_to_index.get)
    base_label = sorted_labels[0]  
    constraints = ({'type': 'eq', 'fun': lambda xi: xi[label_to_index[base_label]]})  
    
    # Minimize the negative log likelihood
    result = minimize(bt_log_likelihood, initial_xi, args=(matches, label_to_index), constraints=constraints)
    
    if result.success:
        return result.x  # Returns the estimated BT coefficients
    else:
        raise ValueError("Optimization failed:", result.message)
\end{lstlisting}
\caption{Estimate Bradley-Terry coefficients for model prediction.}
\label{listing:3}
\end{listing}

\begin{listing}[!ht]
\hrule
\begin{lstlisting}[
language=Python,
basicstyle=\footnotesize\ttfamily,
numbers=left,
breaklines=true,
breakatwhitespace=true,
xleftmargin=2em,
xrightmargin=2em,
aboveskip=1em,
belowskip=1em
]
def rank_competitors(bt_coefficients, label_to_index):
    # Reverse the label_to_index mapping to get labels from indices
    index_to_label = {index: label for label, index in label_to_index.items()}
    
    # Create a list of (label, coefficient) tuples
    labeled_coefficients = [(index_to_label[i], coeff) for i, coeff in enumerate(bt_coefficients)]
    
    # Sort competitors based on coefficients in descending order
    sorted_competitors = sorted(labeled_coefficients, key=lambda x: x[1], reverse=True)
    
    # Format the ranking into a readable format
    rankings = [f"{label}: {coeff:.3f}" for label, coeff in sorted_competitors]
    return rankings
\end{lstlisting}
\caption{Rank competitors based on Bradley-Terry model coefficients.}
\label{listing:4}
\end{listing}

\newpage
\clearpage

\section{Levels of Autonomy in Reviewing}
\label{sec:autonomy_levels}
We currently do not want to fully replace human reviews and their evaluation by AI. Problems with solely using LLMs include evaluation bias, the risk of LLMs favoring results from similar LLMs, potential bias against specific user groups, misinformation, and hallucinations. Our goal is to avoid such biases and ensure factual accuracy. We propose combining humans and LLMs for evaluation by understanding the broad spectrum between human evaluation and full automation by LLMs. Currently, humans are the sole reviewers without any AI interference. It is common practice for humans to maintain complete control while being supported by AI which summarizes and highlights paper and review texts. LLMs may help humans make decisions by generating summaries for human evaluation. Beyond summaries, humans and LLMs may collaborate by having each make decisions they are good at, such as by role-playing and dialogue.
Moving closer to automation is achieved by humans in the loop, having humans prefer between automated evaluations or decisions or having a human verify and accept or reject an automated decision. Role-playing and dialogue may consider LLMs as crowd workers to be supervised by humans. Finally, fully automated reviewing may be beneficial in automating the entire scientific process, but it is unsuitable for the current academic review process.

\end{document}